\documentclass[lettersize,journal]{IEEEtran}
\usepackage{amsmath,amsfonts}
\usepackage[table,xcdraw]{xcolor}
\usepackage{breqn}
\usepackage{algorithmic}
\usepackage[switch]{lineno}
\usepackage{algorithm}
\usepackage{array}
\usepackage{hyperref}
\hypersetup{
    colorlinks=true,
}
\usepackage{multirow}
\usepackage{booktabs}
\usepackage{siunitx}
\usepackage{longtable}
\usepackage{subcaption} % luis added
\usepackage{textcomp}
\usepackage{stfloats}
\usepackage{verbatim}
\usepackage{graphicx}
\usepackage{amstext} % for \text macro
\usepackage{array}   % for \newcolumntype macro
\usepackage{todonotes}

\newcolumntype{L}{>{$}l<{$}} % math-mode version of "l" column type
\def\G{\mathcal{G}}
\def\N{\mathcal{N}}
\def\E{\mathcal{E}}
\def\L{\mathcal{L}}
\def\S{\mathcal{S}}
\def\R{\mathcal{R}}

\definecolor{lightgray}{gray}{0.9}

\begin{document}

\title{CANOS: A Fast and Scalable Neural AC-OPF Solver Robust To N-1 Perturbations}

\author{Luis Piloto, Sofia Liguori\textsuperscript{*}, Sephora Madjiheurem\textsuperscript{*}, Miha Zgubi\v{c}\textsuperscript{*}, Sean Lovett, Hamish Tomlinson, Sophie Elster, Chris Apps\textsuperscript{$\ddag$} and Sims Witherspoon\textsuperscript{$\ddag$}
\thanks{\copyright \hspace{1pt} Google DeepMind. All rights reserved}}
% The paper headers
% \markboth{Journal of \LaTeX\ Class Files,~Vol.~14, No.~8, August~2021}%
% {Shell \MakeLowercase{\textit{et al.}}: A Sample Article Using IEEEtran.cls for IEEE Journals}

% \IEEEpubid{0000--0000/00\$00.00~\copyright~2021 IEEE}
% Remember, if you use this you must call \IEEEpubidadjcol in the second
% column for its text to clear the IEEEpubid mark.

\maketitle

\def\thefootnote{*, \ddag }\footnotetext{\hspace{1pt}These authors contributed equally to this work}\def\thefootnote{\arabic{footnote}}

\begin{abstract}
Optimal Power Flow (OPF) refers to a wide range of related optimization problems with the goal of operating power systems efficiently and securely. In the simplest setting, OPF determines how much power to generate in order to minimize costs while meeting demand for power and satisfying physical and operational constraints. 
In even the simplest case, power grid operators use approximations of the AC-OPF problem because solving the exact problem is prohibitively slow with state-of-the-art solvers. These approximations sacrifice accuracy and operational feasibility in favor of speed. 
This trade-off leads to costly ``uplift payments" and increased carbon emissions, especially for large power grids.
In the present work, we train a deep learning system (\textsc{CANOS}) to predict near-optimal solutions (within 1\% of the true AC-OPF cost) without compromising speed (running in as little as 33--65 ms). 
Importantly, \textsc{CANOS} scales to realistic grid sizes with promising empirical results on grids containing as many as 10,000 buses. 
Finally, because \textsc{CANOS} is a Graph Neural Network, it is robust to changes in topology. We show that \textsc{CANOS} is accurate across N-1 topological perturbations of a base grid typically used in security-constrained analysis. 
This paves the way for more efficient optimization of more complex OPF problems which alter grid connectivity such as unit commitment, topology optimization and security-constrained OPF.
\end{abstract}


%\linenumbers

\section{Introduction}
Electricity is delivered from producers to consumers through interconnected networks known as power grids, consisting of large numbers of components interacting through electrical transmission lines. Power is produced at various types of generators, and needs to be transmitted to consumers ranging from heavy industry to residential users. Grid operators are challenged with the task of ensuring that enough power is reliably and economically dispatched across the network to satisfy demand. It is imperative that this is done while maintaining a balance between power injections and withdrawals and satisfying physical constraints, in order to prevent risks such as overloading of equipment, network outages or failure to supply power to consumers. This requires solving a constrained non-linear non-convex optimization problem known as alternating current optimal power flow (AC-OPF).
Specifically, solving an AC-OPF problem equates to finding a steady state operating point in a power grid that minimizes the cost of power generation while satisfying operating constraints and meeting power demand.
The AC-OPF problem is a cornerstone for a wide range of power systems operations such as unit commitment, security constrained economic dispatch, congestion management and optimal transmission switching.  
In practice, because the demand, supply and grid specification can vary rapidly over time, the AC-OPF problem should ideally be solved close to real-time and many times throughout the day (every five to fifteen minutes).

Relying on traditional optimization tools to solve such a problem on a real-size grid is often computationally infeasible or unachievable in real-time, which has lead to a range of techniques designed to meet grid operators' need for practical solutions to the AC-OPF problem at rapid timescales \cite{Cain2012HistoryOO}. A common approach is to linearize the problem, resulting in the so-called DC-OPF \cite{dc_opf2009} formulation. This formulation is much less computationally expensive, and scalable methods exist \cite{sun2006}. However, solutions to the DC-OPF problem are known to be infeasible for AC-OPF \cite{Baker2021}, and can require post-processing in order to make the solutions of practical use. Even then, these solutions are suboptimal due to the approximation of the original problem. In security-constrained economic dispatch, this means market prices do not correctly represent the true system costs \cite{Sauer2014}, potentially leading to higher carbon emissions \cite{Winner2023} and tens of billions of dollars annually in excess costs \cite{Cain2012HistoryOO}.
More advanced approximation techniques such as convexification \cite{Lavaei2010} or implicit linearization \cite{Bolognani2015}, though closer to the original formulation, are impacted by similar limitations in which speed of computation is favored over accuracy. Even in cases where such approximate solutions are within the feasible region of the original AC-OPF formulation, their suboptimality results in higher costs and increased carbon emissions.
Finding solutions that are close to the AC-OPF solution in near real-time remains an open challenge.

In the past few years, machine learning (ML) methods have emerged as powerful tools for approximating solutions to constrained optimization problems \cite{Kotary2021}. Indeed, in the presence of extensive data, ML techniques, and specifically deep learning methods, offer an alternative to traditional approximate solvers.
Consequently, previous efforts have proposed deep learning based approaches to power systems problems \cite{Khaloie2024}.
Despite their capabilities, ML approaches to constrained optimization problems face their own challenges, since enforcing constraint satisfaction in an end-to-end ML model is non-trivial.

Using the fact that some variables in the AC-OPF formulation can be derived from other independent variables using power flow equations, 
works such as \cite{Zamzam2020} and \cite{Pan2021} train a deep neural network (DNN) to first approximate only the set of independent decision variables, then derive the remaining ones.
The constraint satisfaction problem has been approached with techniques such as combining Lagrangian methods with deep learning \cite{Fioretto2020}, and adopting a physics-guided graph convolution neural network which embeds both physical features and operational constraints \cite{Gao2024}.

In addition to constraint satisfaction, adaptability to topology variations is critical.
In practice, power grids experience topological variations stemming from unpredictable short-term outages (contingencies), planned maintenance, or medium-term development of the grid. As a result, there is a need for ML models that can reliably predict solutions under topological perturbations \cite{Popli2024}.
Recent work in this direction shows promise, but results on AC-OPF with topological perturbation are generally limited to smaller grids, or do not report constraint satisfaction metrics \cite{Chen2022,Falconer2023,Gao2024,Liu2023,Popli2024,Nakiganda2023,Yang2023}.
An exception is the work of \cite{Zhou2023}, which reports results for N-1 contingency experiments for a large 2,000-bus system, including mean percentage of constraints satisfied. However changes in topology are represented by setting line admittance features to zero, which means that the resulting trained models can only drop lines from the full specification; they cannot easily handle general perturbations such as adding arbitrary new lines.

In our work, we present a solver that can approximate \textit{near-optimal}, \textit{near-feasible}, and \textit{robust} solutions \textit{efficiently}. The main contributions of the present work can be summarized as follows:
\begin{enumerate}
    \item We propose \textsc{CANOS} (Constraint-Augmented Neural OPF Solver), a method based on Graph Neural Networks (GNNs) for learning a fast, accurate, scalable, and robust AC-OPF solver.
    \item We empirically demonstrate \textsc{CANOS}' ability to generate accurate solutions (within an optimality gap of $1\%$) and extensively document constraint violation metrics for all relevant constraints.
    \item We show that \textsc{CANOS} scales to realistic grid sizes containing as many as 10,000 buses, and remains robust under typical topological perturbations used in security-constrained analysis.
    \item We empirically show that \textsc{CANOS} outperforms DC approximations in terms of accuracy and AC-feasibility.
    \item We demonstrate that \textsc{CANOS} is fast, running in 33-65 ms without power flow post-processing for grids between 500--10,000 buses, showing sub-linear scaling with grid size.
\end{enumerate}
To the best of the authors' knowledge this work is the first to demonstrate a successful application of GNNs to solving AC-OPF on large (10,000-bus) grids with extensively documented feasibility analysis of the solutions and a fully-general handling of topological perturbations.
Our model leverages GNNs to handle different topologies without relying on zeroing-out features of existing entities, which means the approach can potentially handle addition of entities and account for more general topological variations that can arise in contexts different from contingency analysis.

The remainder of this paper is structured as follows. Section~\ref{sec:formulation} introduces the notation used throughout the manuscript and describes the problem formulation, while Section~\ref{sec:data} describes the data used for training and evaluating our model. In  Section~\ref{sec:canos}, we give a brief introduction to the deep learning architecture used in this work, and describe our proposed model in detail. Sections~\ref{sec:training} and \ref{sec:eval} present the details of our training and evaluation procedures.  
In Section~\ref{sec:results} we present, analyse and discuss our empirical findings. Finally, we conclude in Section~\ref{sec:conclusion}, discussing the limitations of our approach and avenues for future work. Appendices~\ref{apdx:data}-\ref{apdx:model_size_analysis} gather supplementary material. 

\section{Problem Formulation}
\label{sec:formulation}
In this section, we introduce the notation and the AC-OPF problem formulation used in this work. The formulation refers to the baseline solver \texttt{PowerModels.jl} \cite{powermodels}.

Consider a power grid consisting of $N$ buses denoted by set $\N$, a set of branches (either AC-lines or transformers) $\E$ and their corresponding reverse orientation $\E^R$, a set of generators $\G$, a set of loads $\L$ and a set of shunts $\S$. We denote by $\R$ the subset of reference buses ($\R \subset \N$).  Generators, loads and shunts are located at buses, and we use subscripts to denote their corresponding bus (e.g. $\G_i$ denotes generators at bus $i$). Furthermore, we adopt the following notation to describe the relevant values in the power grid:
\vspace{10px}
\\
\vspace{5px}
\begin{tabular}{L L}
S_k^g \,\, \forall k \in \G  & \small\text{Generator complex power dispatch}\\[3pt]
S_k^{gl}, S_k^{gu} \,\, \forall k \in \G &\small\text{Generator complex power bounds}\\[3pt]
c_{2k}, c_{1k}, c_{0k} \,\, \forall k \in \G &\small\text{Generator cost components}\\[3pt]
V_i \,\, \forall i \in \N & \small\text{Bus complex voltage}\\[3pt]
v_i^l, v_i^u \,\, \forall i \in \N &\small\text{Bus voltage magnitude bounds} \\[3pt]
S_{ij} \,\, \forall (i, j) & \small\text{Branch complex power flow}\\
S_k^{d} \,\, \forall k \in \L &\small\text{Load complex power consumption} \\[3pt]
Y_k^{s} \,\, \forall k \in \S &\small\text{Bus shunt admittance} \\[3pt]
Y_{ij}, Y_{ij}^c, Y_{ji}^c \,\, \forall (i, j) \in \E &\small\text{Branch pi-section parameters} \\ [3pt]
T_{ij} \,\, \forall (i, j) \in \E &\small\text{Branch complex transformation} \\
& \small\text{ratio} \\[3pt]
s_{ij}^u \,\, \forall (i, j) \in \E &\small\text{Branch apparent power limit} \\[3pt]
\theta_{ij}^{\Delta l}, \theta_{ji}^{\Delta u} \,\, \forall (i, j) \in \E &\small\text{Branch voltage angle difference} \\
& \small\text{bounds}  
\end{tabular}
where the superscripts $l$ and $u$ respectively indicate lower and upper bounds. 

The complete mathematical formulation is defined in terms of these quantities as follows\footnote{ $\Re(X)$ denotes the real coefficient of a complex variable $X$, $\Im(X)$ corresponds to its imaginary coefficient, $\angle X$ represent the complex angle, and $X^*$ denotes the complex conjugate of $X$}:
\begin{align}
\nonumber \text{variables:} & \\
& S_k^g \,\, \forall k \in \G   \label{eq:gen_pow} \\
& V_i \,\, \forall i \in \N   \label{eq:bus_vol} \\
& S_{ij} \,\, \forall (i, j) \in \E \cup \E^R  \label{eq:branch_pf} \\
\nonumber \text{minimize:} & \\
& \sum_{k\in G}c_{2k} \Re(S_k^g)^2 + c_{1k}\Re(S_k^g) \label{eq:cost}
\end{align}
\begin{align}
 \nonumber \text{subject to:}  \\
& \angle V_r = 0 \,\, \forall r \in \R \label{eq:constr1}\\
& S_k^{gl} \leq S_k^{g}  \leq S_k^{gu} \,\, \forall k \in \G \label{eq:constr2}\\
& v_i^l \leq | V_i | \leq v_i^u \,\, \forall i \in \N \label{eq:constr3}\\
\nonumber & \sum_{k\in\G_i} S_k^g - \sum_{k\in\L_i} S_k^d - \sum_{k\in\S_i} (Y_k^s)^*| V_i|^2 \label{eq:constr4}\\ 
& = \sum_{(i,j) \in \E_i \cup \E_i^R } S_{ij} \,\, \forall i \in \N \\
& S_{ij} = \big( Y_{ij} + Y_{ij}^c \big)^* \frac{|V_i|^2}{|T_{ij}|^2} - Y_{ij}^* \frac{V_i V_j^*}{T_{ij}} \,\, \forall (i,j) \in \E \label{eq:constr5}\\
& S_{ji} = \big( Y_{ij} + Y_{ij}^c \big)^* |V_j|^2 - Y_{ij}^* \frac{V_i^* V_j}{T_{ij}} \,\, \forall (i,j) \in \E \label{eq:constr6}\\
& |S_{ij}| \leq s^u_{ij} \,\, \forall (i,j) \in \mathcal{E} \cup \mathcal{E}^R \label{eq:constr7}\\
& \theta^{\Delta l}_{ij} \leq \angle (V_i V^*_j) \leq \theta^{\Delta u}_{ij} \;\; \forall (i,j) \in \mathcal{E} \label{eq:constr8}
\end{align}

\begin{itemize}
    \item Equation~\eqref{eq:constr1} enforces the reference bus voltage angle to be 0.
    \item Inequalities~\eqref{eq:constr2} are the \textbf{generator power bounds}, limiting the capacity of generators in the equipment physical range.
    \item Inequalities~\eqref{eq:constr3} are the \textbf{voltage magnitude bounds}.
    \item Equation~\eqref{eq:constr4} is the \textbf{bus power balance} equation, enforcing that the net injected power at each bus equals the power flowing through the branches connected to the bus.
    \item Equation~\eqref{eq:constr5} represents the power flowing in a branch in the ``from" direction: from bus $i$ to bus $j$, while eq.~\eqref{eq:constr6} represents the power flowing in the opposite direction, the ``to" direction, from bus $j$ to bus $i$. We collectively refer to these equations as \textbf{branch flow} equations or as \textbf{Ohm's constraints}.
    \item Inequality~\eqref{eq:constr7} is the \textbf{branch thermal limit}, which represents the maximum power that can flow in a branch without it being damaged by overheating.
    \item Inequalities~\eqref{eq:constr8} are bounds on the \textbf{voltage angle difference} of connected buses, which guarantees stability of the network.
\end{itemize}

\section{Dataset}
\label{sec:data}

We build our training and evaluation datasets starting from the base test cases of Power Grid Library (\textsc{PGLib-OPF}) \cite{pglib}, a widely-used AC-OPF benchmark. Each example in \textsc{PGLib-OPF} contains data about the grid specification and power demand. We detail the data format in Appendix~\ref{apdx:data}. 

We use \texttt{PowerModels.jl} \cite{powermodels}, a network optimization package, and the nonlinear optimization solver \texttt{Ipopt} \cite{ipopt} to generate the AC-OPF solution. The solutions are organised in graphs of the following format, with all power variables expressed in the per-unit system:\\
\nopagebreak
\textbf{Solution nodes} \vspace*{4px}\\
\begin{tabular}{l l}
\textbf{Bus} &  \\
\texttt{va} ($\angle (V_i V^*_j)$) & \small{Voltage angle}\\
\texttt{vm} ($|V_i|$) & \small{Voltage magnitude}\\
\textbf{Generator} &  \\
\texttt{pg} ($\Re(S^g_k)$) & \small{Real power generation}\\
\texttt{qg} ($\Im(S^g_k)$) & \small{Reactive power generation}\\
\end{tabular}\\
\vspace*{1px}\\
\textbf{Solution edges} \vspace*{4px} \nopagebreak\\
\begin{tabular}{l l}
\textbf{AC-line} &  \\
\texttt{pt} ($\Re(S_{ji})$) & \small{Active power withdrawn at the to bus}\\
\texttt{qt} ($\Im(S_{ji})$) & \small{Reactive power withdrawn at the to bus}\\
\texttt{pf} ($\Re(S_{ij})$) & \small{Active power withdrawn at the from bus}\\
\texttt{qf} ($\Im(S_{ij})$) & \small{Reactive power withdrawn at the from bus}\\
\textbf{Transformer} &  \\
\texttt{pt} ($\Re(S_{ji})$) & \small{Active power withdrawn at the to bus}\\
\texttt{qt} ($\Im(S_{ji})$) & \small{Reactive power withdrawn at the to bus}\\
\texttt{pf} ($\Re(S_{ij})$) & \small{Active power withdrawn at the from bus}\\
\texttt{qf} ($\Im(S_{ij})$) & \small{Reactive power withdrawn at the from bus}\\
\end{tabular}\\
\vspace*{2px}
\\
We build two types of datasets, \textsc{FullTop} and \textsc{TopDrop}, for three different grids: the 500-bus base case, the 2,000-bus base case and the 10,000-bus base case\footnote{See Appendix \ref{apdx:data} for a complete description of the number of grid elements.}. For each grid size, we generate the two datasets, each consisting of 300k examples. We split each dataset randomly into training, validation, and test sets, with 90\%, 5\%, and 5\% of examples, respectively.
\\
\textbf{\textsc{FullTop} dataset:}
Our first dataset type is the full topology dataset, \textsc{FullTop}, which we build from the reference \textsc{PGLib-OPF} example by perturbing each load independently by $20\%$ of its value (both the active and reactive components are independently perturbed) similar to \cite{Fioretto2020,Zamzam2020}, and keeping the topology intact.
\\
\textbf{\textsc{TopDrop} dataset:}
The second type of dataset is the \textsc{TopDrop} dataset. In addition to load profile perturbation, we also perturb the topology of the grid for 50\% of the examples. The topology alteration is a N-1 perturbation achieved by uniformly selecting a generator (excluding generators connected to the slack bus) or uniformly selecting a branch (AC-line or transformer) with equal probability to drop. The dropped components are removed from the grid entirely as long as it does not result in a disconnected graph, in which case we sample a different component.

\section{Constraint Augmented Neural OPF Solver}
\label{sec:canos}

Graph Neural Networks (GNNs) are a specialized type of neural network architecture designed to work directly with structured data. Graphs are composed of a set of nodes representing entities, and a set of edges representing the pairwise relationships between the entities. Each of these graph components have features associated with them. The key feature of GNNs is the concept of \textit{message passing}. 
Message passing allows the learning of a local representation of nodes and edges by iteratively passing messages (a function of node,  edge and global features) along the edges of the graph. These representations can then be used for downstream tasks such as node and edge prediction \cite{Battaglia2018}. 
This type of neural network is naturally well suited for our problem, as a power grid is intuitively represented as a heterogeneous graph, and the AC-OPF problem can be formulated as a prediction problem on a subset of this graph.

\begin{figure}[!t]
\centering
\includegraphics[width=3.5in]{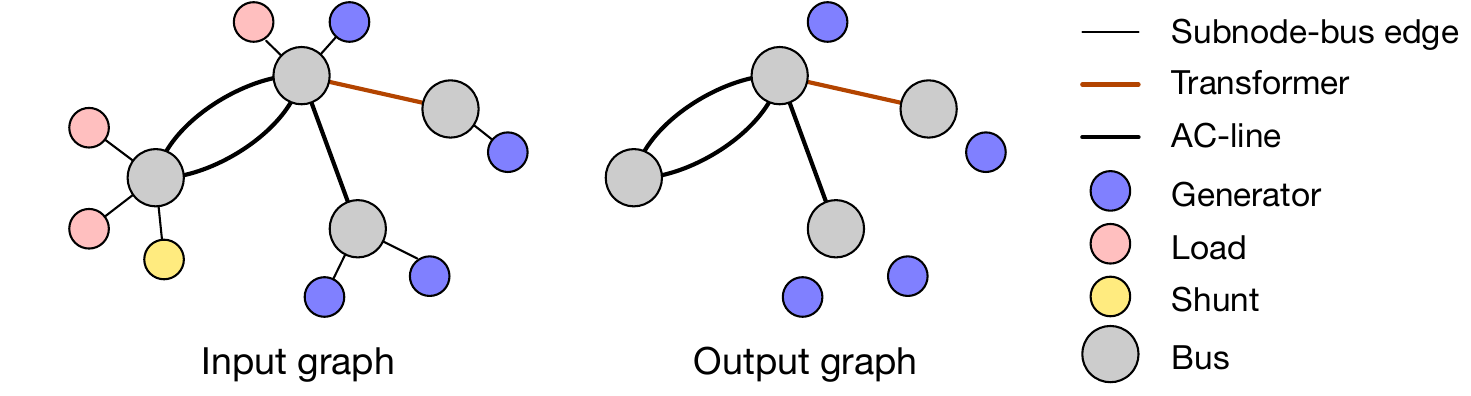}
\caption{\textsc{CANOS} input (left) and output (right) graph structures. Different node and edge types are represented in different colors. The output graph only contains entities with predicted features.}
\label{fig:graph_data}
\end{figure}

More concretely, the nodes in the power grid graph are the $N$ buses and their corresponding ``subnodes'' (generators, loads and shunts) and the edges are the transformers and AC-lines (represented as different types of edges). We also use artificial edges to connect the subnodes to their respective bus (these do not model any physical equipment). We model branches with directed edges, and ensure to propagate edge features to nodes in both directions during the message passing step.
Figure~\ref{fig:graph_data} (left) shows an illustration of a grid represented as a graph (which we refer to as \textit{input graph}). Each node and edge in the input graph has a feature vector containing the corresponding values described in Appendix~\ref{apdx:data}. 
The quantities to be predicted in the AC-OPF formulation (\eqref{eq:gen_pow}-\eqref{eq:branch_pf}) can all be represented as signals on a subgraph (the graph with shunts, loads, and artificial edges removed). The node and edge features of the \textit{output graph}  are described by the solution nodes and solutions edges in Section~\ref{sec:data}. The same graph structure is shared by the model output graph and the target graph representing the dataset solution.
We standardize input features and targets to 0 mean and standard deviation 1 across the training set.
Figure~\ref{fig:graph_data} (right) shows the output/target graph for the example on the left.

In the following, we describe in detail the Constraint-Augmented Neural OPF Solver (\textsc{CANOS}), our GNN based AC-OPF solver. 

\begin{figure*}[!t]
\centering
\includegraphics[width=\textwidth]{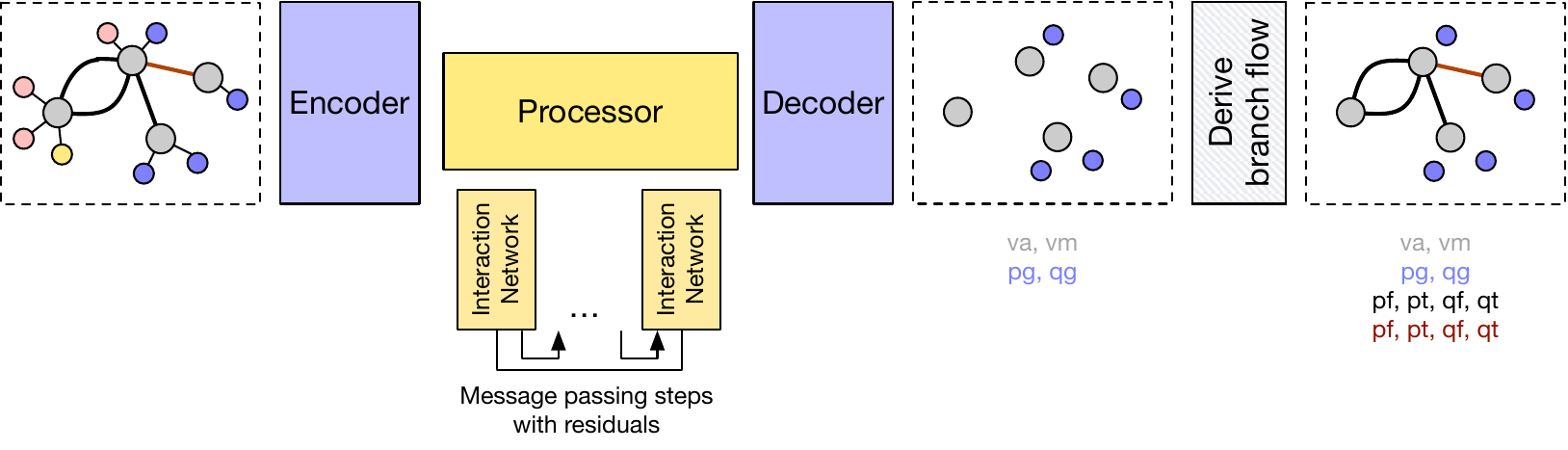}
\caption{\textsc{CANOS} architecture. The colored layers have trainable weights, while the striped module is a non-parameterized graph function. The decoder outputs the voltage angle $\texttt{va} = \angle (V_i V^*_j)$ and voltage magnitude $\texttt{vm} = |V_i|$ as bus node features, as well as quantities relating to generator dispatched power (real power $\texttt{pg} = \Re(S^g_k)$ and reactive power $\texttt{qg} =\Im(S^g_k)$) as generator node features. The branch flow derivation module uses these predicted values to compute the branch complex power in the two directions (real and reactive powers  $\texttt{pf}$, $\texttt{qf}$, $\texttt{pt}$, $\texttt{qt}$) according to Equations~\eqref{eq:constr5} and \eqref{eq:constr6}.}
\label{fig:architecture}
\end{figure*}

\textsc{CANOS} (depicted in Figure~\ref{fig:architecture}) is an encode-process-decode architecture \cite{Hamrick2018, Battaglia2018}, followed by a branch flow derivation module. Each element of the architecture is composed of one or more graph network (GN) blocks. GNs are graph-to-graph modules which perform a sequence of computations over the structure. See \cite{Battaglia2018} for a detailed definition of the internal structure of a GN. 
\\
\textbf{Encode:} The features of the input graph are independently projected onto a latent representation. The encoder module is a GN block in which the edge and node update functions are linear networks which project all node and edge features into a vector of size \texttt{hidden\_size}. \\
\textbf{Process:}
The processor consists of a sequence of \texttt{num_message_passing_steps} interaction network (IN) blocks \cite{Battaglia2016} with residual connections, where the edge and node update functions are implemented as 2-layer MLPs of size \texttt{hidden_size}, with relu activation and layer normalization. We adopt the IN augmentation introduced in \cite{Allen2023} to enable message passing over nodes and edges of different types - where each different type of node or edge has an update function with unique parameters. See Table \ref{tab:models} for the number of message passing steps used in different variants of CANOS.\\
\textbf{Decode:} The processed latent features are then decoded into output values for bus voltage and generator power for all buses and generators. The decoder is a GN block with edge and node update functions implemented as a 2-layer MLP of size 256, with relu activation and layer normalization, followed by a linear layer projecting the latent features into output vectors.
We enforce bound constraints \eqref{eq:constr2}--\eqref{eq:constr3} on the corresponding outputs $y$ using the sigmoid function:
\begin{equation}
y = \text{sigmoid}(y)*(y^u - y^l) + y^l
\label{eq:sigmoid_bound}
\end{equation}

\textbf{Derive branch flow:} The output of the decoder is then used to derive the remaining variables (branch power) according to the branch flow equations \eqref{eq:constr5} and \eqref{eq:constr6}.

We adopt variants of the model architecture, shown in Table~\ref{tab:models}.

\begin{table}[ht]
\centering
\begin{tabular}{|l|c|c|}
\hline
 & message passing steps & hidden size \\ \hline
\textsc{deep-CANOS}$_{48}$ & 48 & 128 \\ \hline
\textsc{deep-CANOS}$_{60}$ & 60 & 128 \\ \hline
\textsc{wide-CANOS}     & 36 & 384 \\ \hline
\end{tabular}
\caption{\textsc{CANOS} variants. We only show hyperparameters that vary across our models.}
\label{tab:models}
\end{table}

\section{Training}
\label{sec:training}
We train \textsc{CANOS} to simultaneously minimize the $L_2$ loss between the prediction and the ground truth as well as minimize constraint violations (\eqref{eq:constr1}--\eqref{eq:constr6}) (constraint augmentation of the loss) along similar lines as \cite{Fioretto2020}. Formally, our loss function consists of two terms:
\begin{equation}
    \ell (G; \boldsymbol w) :=  \, \, \ell_{\text{supervised}}(G; \boldsymbol w) + C \ell_{\text{constraints}}(G; \boldsymbol w) \,,
\end{equation}
where $G$ designates the input graph, $\boldsymbol w$ represents the GNN parameters, and $C$ is a constraint weighing coefficient. We set $ C = 0.1 $ across all CANOS variants. The supervised loss $\ell_{\text{supervised}}$ aggregates the $L_2$ losses for the bus voltage, the generator power, and the branch power between the predicted values and the targets. The constraint loss $\ell_{\text{constraints}}$ aggregates the \textit{constraint violation degrees} \cite{Fioretto2020} for each constraint. For equality constraints, a constraint degree measures the absolute distance between the left-hand and right-hand sides of the equations. For inequality constraints, this measures the distance to the closest boundary (or equals zero when the constraint is satisfied). Finally, note that we split complex-valued constraints e.g. Equation~\eqref{eq:constr4} into real and reactive components.

We train our GNN modules with gradient descent with respect to the weights $\boldsymbol w$, using the Adam optimizer. We train models for 600,000 steps, except for the models running the 10,000-bus grids which are trained between 600,000 and 800,000 steps. We train on 8 A100 40GB GPUs using a global batch size of 32 for the 500-bus grid datasets and 16 for the larger grid sizes.

For the first 10,000 steps, we ramp up the learning rate from $0$ to $2\times10^{-4}$ following a linear schedule. Then, we use an exponential decay schedule with decay rate $0.9$ and transition steps $4000$\footnote{Transition steps define the granularity of the schedule's exponent update. See \texttt{Optax} exponential decay schedule for more details.} to reach the value of $5\times10^{-6}$.

We implement our models, training and evaluation pipeline using \texttt{JAX} \cite{jax2018github}, \texttt{haiku} \cite{haiku2020github}, \texttt{jraph} \cite{jraph2020github}, \texttt{jaxline} \cite{deepmind2020jax}, and \texttt{Optax} \cite{deepmind2020jax}.
 
\section{Evaluation}
\label{sec:eval}
We perform an extensive evaluation of \textsc{CANOS} to assess the suitability of the proposed method in real-world AC-OPF challenges. After training, we report metrics using the test split of the datasets consisting of roughly 15,000 examples for each combination of grid size (500-, 2,000-, 10,000-bus) and dataset condition (\textsc{FullTop}, \textsc{TopDrop}). We train three \textsc{CANOS} models using different random seeds. We report metrics covering speed, optimality, solution accuracy, and feasibility as the average across those three seeds. We assess our model against two baseline methods: a full AC-OPF solver and an approximate DC-OPF solver. We also report our results before and after post-processing with AC power flow. This reflects the two ways that approximate solvers, like \textsc{CANOS} and DC-OPF, are used. Some applications do not require approximate solutions to strictly satisfy AC constraints (e.g. linear relaxations are frequently used in stochastic OPF \cite{yong2000stochastic} or mixed-integer problems such as unit commitment \cite{van2014dc}). 
In other applications, approximate solutions are used with the power flow post-processing step to ensure AC-feasibility in an iterative refining process\cite{ONeill2011}.

Our thorough analysis before and after power flow addresses suitability for both use cases.

\subsection{Baselines}
We compare \textsc{CANOS} against the performance of a full AC-OPF solver (\textsc{AC-IPOPT}) and an approximate DC-OPF solver (\textsc{DC-IPOPT}). \textsc{AC-IPOPT} solutions are the ``gold standard'' in terms of solution quality, but can be prohibitively slow to compute \footnote{They are also not guaranteed to converge or find global minima \cite{Cain2012HistoryOO}.}. On the other hand, \textsc{DC-IPOPT} is much quicker to run. It solves the DC-OPF formulation \cite{dc_opf2009} which is only an \textit{approximation} to the full AC-OPF problem. Importantly, the DC-OPF solutions are not feasible with respect to the full set of AC-OPF constraints \cite{Baker2021}. Furthermore, DC-OPF only provides a \textit{partial} solution to the AC-OPF problem, as it does not specify the reactive power components. Despite these shortcomings, the speed-up endowed by DC-OPF solvers is compelling enough that they are widely used in various grid optimization settings across network planning and operation \cite{ONeill2011,Mancarella2020}. Thus, we examine our model's performance against these two bookend baselines which span the spectrum from ``slow and feasible'' to ``fast and approximate.''

\begin{enumerate}
\item \textbf{AC baseline} \textsc{AC-IPOPT}: The reference solution is the full AC-OPF solution generated using the open source Julia solver {\texttt{PowerModels.jl}}\cite{powermodels} with the \texttt{Ipopt} optimizer.  
    
\item \textbf{DC baseline} \textsc{DC-IPOPT}:  The approximate solution is the DC-OPF solution also generated using the open source Julia solver {\texttt{PowerModels.jl}}\cite{powermodels} with the \texttt{Ipopt} optimizer.

\end{enumerate}

\subsection{Running Power Flow}
We use the open source library \texttt{pandapower} \cite{pandapower} with numba acceleration to run AC-power flow starting from either the \textsc{DC-IPOPT} solution or \textsc{CANOS} AC-OPF solution.
We use the option \texttt{enforce\_q\_lims=True} to guarantee that the generator reactive power stays within bounds after power flow (slack excluded).
The new values \texttt{va'}, \texttt{vm'}, \texttt{pg'}, \texttt{qg'}, \texttt{pf'}, \texttt{qf'}, \texttt{pt'}, \texttt{qt'} constitute the \textit{post-power flow solution} and are used to compute \textit{post-power flow metrics}. Power flow guarantees that the  power balance Equation ~\eqref{eq:constr4} is satisfied, but other constraints can still be violated (see Appendix \ref{apdx:power_flow} for more details on power flow and its implications on constraints violations). Note also that after AC-power flow, the \textsc{DC-IPOPT} solution, which originally lacks all reactive components, becomes a complete AC solution.

\subsection{Metrics}

\subsubsection{Supervised Metrics}
Supervised metrics
measure \textit{solution fidelity}:  how closely a solution matches the ``gold standard'' provided by the AC-OPF solver. Evaluating solution fidelity is important for two different reasons.

First, having accurate solutions is particularly critical in the applications that require approximate solvers.
For example in year-long network planning, transmission system operators (TSOs) often use DC-OPF to solve a sequence of OPF problems for every hour of a single week. Because there are so many problems to solve, a fast, approximate solver must be used. However, solutions earlier in the week condition solutions later in the week. Hence, errors accumulate on successive dispatch decisions and accuracy becomes critical.

Second, providing supervised metrics allows for comparison of the efficacy of various ML-for-OPF models, the effect of hyperparameters and different training or optimization procedures.

To quantify supervised performance, we use the thresholded relative mean-absolute error (TRMAE). The TRMAE computes the absolute error and reports it relative to the magnitude of the target solution from AC-OPF. We consider the \textsc{AC-IPOPT} solution as the ground truth target. This is computed for each feature of the OPF solution. We apply a threshold of $0.001$ to avoid explosions in this relative metric due to small values in the targets. This threshold removes from the mean those values with small targets from both the \textsc{DC-IPOPT} and \textsc{CANOS} bars - thus allowing for a fair comparison between the two methods.
We report the TRMAE because it's a more intuitive scale to understand errors, but see Appendix  \ref{apdx:additional_supervised_metrics} for results using mean-squared error (MSE), which includes all examples and is important for comparison with other ML models. 

\subsubsection{AC Feasibility - Before Power Flow}

The AC-feasibility of a solution is determined by whether or not it satisfies the AC constraints, Equations~\eqref{eq:constr1}--\eqref{eq:constr8}. Henceforward we refer to this simply as ``feasibility.'' We can evaluate the feasibility before and after post-processing with power flow.

When we evaluate the feasibility of \textsc{CANOS}'s OPF solutions before power flow, we assess whether or not any post-processing is needed at all. This, of course, depends on the specific application and constraint satisfaction thresholds required.

To quantify feasibility, we evaluate the degree violations (the same used in the training loss) for all constraints which are not \textit{guaranteed} to be $0$.

\textsc{CANOS} guarantees satisfaction of:
\begin{itemize}
    \item The reference bus angle constraint, Equation~\eqref{eq:constr1}.
    \item The power and voltage bounds, Equations~\eqref{eq:constr2} and \eqref{eq:constr3}, which we enforce with the sigmoid function \eqref{eq:sigmoid_bound}.
    \item The branch flow equations (ohm constraints), Equations~\eqref{eq:constr5}--\eqref{eq:constr6}, which are guaranteed to be satisfied because we derive the branch flow outputs.
\end{itemize}
We quantify the violation for all remaining constraints (power balance, voltage angle difference, thermal limits). Finally, note that because \textsc{DC-IPOPT} only provides a partial solution, we don't evaluate its AC feasibility before the power flow step.

In Appendix~\ref{apdx:additional_supervised_metrics} we also report the percentage of entities satisfying constraints at different threshold levels. This information complements the reported average violations to give a better understanding of the distribution of such violations across entities.

\subsubsection{AC Feasibility - After Power Flow}

Solving the power flow equations guarantees that the bus power balance constraints are satisfied. However, other constraints can be violated as the power flow procedure adjusts bus voltages, reactive power, and slack generator power. 
Hence, the subset of post-power flow non-zero constraints is different from the pre-power flow one.
Furthermore, some constraints are only violable depending on the bus type. For instance, the real generator power bounds can only be violated for generators connected to the slack bus. Thus, in some instances, we only report constraints by the appropriate type of unit - otherwise averaging over the full set of units would artificially decrease violations. Refer to Appendix \ref{apdx:power_flow} for a
comprehensive description of the implications of power flow on constraint violations.

\subsubsection{Optimality}

The objective value of an AC-OPF solution measures optimality, representing the financial cost (or benefit) of a dispatch decision. Because the objective value depends only on the real power, we can evaluate it on both full AC-OPF solutions, as well as partial DC-OPF solutions. Thus, we compare the optimality of \textsc{CANOS} and \textsc{DC-IPOPT} against the reference (ground truth) objective value provided by \textsc{AC-IPOPT} before and after power flow.
We report the ratio:
\begin{equation}
\text{Optimality} = 
\frac{\text{Objective}(\text{solution})}{\text{Objective}(\text{AC-IPOPT})}
\end{equation}
averaged over the dataset.

\subsubsection{Speed}
Although the AC-OPF formulation only defines optimality and feasibility metrics, a solver's speed is an equally important metric. If a solver takes too long to produce a solution then it is fundamentally blocked from certain applications. For example, an AC-OPF solver that takes 6 minutes to produce a solution cannot be used for dispatch decisions made every 5 minutes. Other applications don't have strict time limits, but they require solving \textit{many} OPF problems.  If a solver takes too long on a single problem, then it cannot be used for these applications. Thus, we evaluate \textsc{CANOS}'s speed to find where it is a viable substitute to DC-OPF and AC-OPF.

We highlight an \textbf{important caveat} here: the processing times for \textsc{AC-IPOPT} and \textsc{DC-IPOPT} were generated with \texttt{PowerModels.jl} using the default, open-source MUMPS \cite{amestoy2000mumps} linear solver in \texttt{Ipopt}. Furthermore, the baselines were run on bulk cloud compute with varying hardware specifications. We expect improvements for both of these baselines if we switched to other solvers such as \texttt{HSL ma57} \cite{duff2004ma57} or potentially ran it on dedicated hardware.  Improving the underlying linear solver would also improve timings for \textsc{CANOS} because the bulk of its processing time comes from the power flow step, which leverages a linear solver, too.  Before power flow, the processing times we see from \textsc{CANOS} are comparable to other deep learning based approaches.  For example, in \cite{Fioretto2020} they report a runtime of 1 millisecond for their OPF-DNN on a 300-bus grid with a $> 10^4$ speedup over the AC-OPF solution. For the 500-bus grid, \textsc{CANOS} runs in 30 milliseconds. This would translate to a speedup of $> 10^2$. This is on the smallest grid we have tested here and we observe sub-linear scaling with grid size which would endow even larger speed-ups. Therefore, we expect similar speed-ups with respect to better baselines as those reported across the ML-for-OPF literature.
\section{Results}
\label{sec:results}

\subsection{Before Post-Processing With Power Flow}

In this section, we evaluate the raw outputs from \textsc{CANOS} and \textsc{DC-IPOPT} before power flow post-processing. These solutions confer the greatest speed-up because they don't require an additional post-processing step. However, they are not guaranteed to satisfy the bus power balance constraints.

\subsubsection{Supervised Metrics}

\begin{table*}[!t]
\centering
\resizebox{\textwidth}{!}{\begin{tabular}{llrrr|rrr|rrr}
\hline
 &  & \multicolumn{3}{c|}{500} & \multicolumn{3}{c|}{2000} & \multicolumn{3}{c}{10000} \\
 &  & \textsc{Wide-CANOS} & \textsc{Deep-CANOS}$_{48}$ & \textsc{DC-IPOPT} & \textsc{Deep-CANOS}$_{48}$ & \textsc{Deep-CANOS}$_{60}$ & \textsc{DC-IPOPT} & \textsc{Deep-CANOS}$_{48}$ & \textsc{Deep-CANOS}$_{60}$ & \textsc{DC-IPOPT} \\
\hline
\multirow[c]{4}{*}{Bus } & \multirow[c]{2}{*}{\texttt{va}} & 1.1\% & \bfseries 0.4\% & 19.9\% & 1.5\% & \bfseries 1.5\% & 22.8\% & 15.4\% & \bfseries 15.3\% & 148.9\% \\
 &  & {\cellcolor{lightgray}} 4.0\% & \bfseries {\cellcolor{lightgray}} 1.2\% & {\cellcolor{lightgray}} 19.9\% & {\cellcolor{lightgray}} 1.9\% & \bfseries {\cellcolor{lightgray}} 1.9\% & {\cellcolor{lightgray}} 22.9\% & \bfseries {\cellcolor{lightgray}} 15.8\% & {\cellcolor{lightgray}} 15.9\% & {\cellcolor{lightgray}} 148.6\% \\
 & \multirow[c]{2}{*}{\texttt{vm}} & 0.0\% & \bfseries 0.0\% & 6.4\% & 0.1\% & \bfseries 0.1\% & 6.1\% & 0.2\% & \bfseries 0.2\% & 6.1\% \\
 &  & {\cellcolor{lightgray}} 0.0\% & \bfseries {\cellcolor{lightgray}} 0.0\% & {\cellcolor{lightgray}} 6.4\% & {\cellcolor{lightgray}} 0.1\% & \bfseries {\cellcolor{lightgray}} 0.1\% & {\cellcolor{lightgray}} 6.1\% & {\cellcolor{lightgray}} 0.2\% & \bfseries {\cellcolor{lightgray}} 0.2\% & {\cellcolor{lightgray}} 6.1\% \\
\multirow[c]{4}{*}{Gen} & \multirow[c]{2}{*}{\texttt{pg}} & 0.3\% & \bfseries 0.1\% & 1.4\% & 0.4\% & \bfseries 0.4\% & 2.8\% & \bfseries 0.4\% & 0.4\% & 2.1\% \\
 &  & {\cellcolor{lightgray}} 0.5\% & \bfseries {\cellcolor{lightgray}} 0.2\% & {\cellcolor{lightgray}} 1.4\% & {\cellcolor{lightgray}} 0.5\% & \bfseries {\cellcolor{lightgray}} 0.5\% & {\cellcolor{lightgray}} 2.8\% & \bfseries {\cellcolor{lightgray}} 0.3\% & {\cellcolor{lightgray}} 0.4\% & {\cellcolor{lightgray}} 2.1\% \\
 & \multirow[c]{2}{*}{\texttt{qg}} & 4.5\% & \bfseries 3.0\% & - & 19.2\% & \bfseries 15.4\% & - & 25.4\% & \bfseries 25.1\% & - \\
 &  & {\cellcolor{lightgray}} 9.9\% & \bfseries {\cellcolor{lightgray}} 7.3\% & {\cellcolor{lightgray}} - & {\cellcolor{lightgray}} 21.2\% & \bfseries {\cellcolor{lightgray}} 17.4\% & {\cellcolor{lightgray}} - & {\cellcolor{lightgray}} 25.8\% & \bfseries {\cellcolor{lightgray}} 25.4\% & {\cellcolor{lightgray}} - \\
\multirow[c]{8}{*}{Line } & \multirow[c]{2}{*}{\texttt{pf}} & 3.2\% & \bfseries 1.5\% & 11.4\% & 5.7\% & \bfseries 4.5\% & 21.1\% & 20.3\% & 21.5\% & \bfseries 18.4\% \\
 &  & {\cellcolor{lightgray}} 6.4\% & \bfseries {\cellcolor{lightgray}} 4.0\% & {\cellcolor{lightgray}} 11.3\% & {\cellcolor{lightgray}} 5.9\% & \bfseries {\cellcolor{lightgray}} 5.1\% & {\cellcolor{lightgray}} 21.2\% & {\cellcolor{lightgray}} 19.6\% & {\cellcolor{lightgray}} 21.3\% & \bfseries {\cellcolor{lightgray}} 18.4\% \\
 & \multirow[c]{2}{*}{\texttt{pt}} & 3.2\% & \bfseries 1.5\% & 11.6\% & 5.7\% & \bfseries 4.5\% & 21.0\% & 20.3\% & 21.5\% & \bfseries 18.4\% \\
 &  & {\cellcolor{lightgray}} 6.3\% & \bfseries {\cellcolor{lightgray}} 4.0\% & {\cellcolor{lightgray}} 11.5\% & {\cellcolor{lightgray}} 5.9\% & \bfseries {\cellcolor{lightgray}} 5.1\% & {\cellcolor{lightgray}} 21.2\% & {\cellcolor{lightgray}} 19.6\% & {\cellcolor{lightgray}} 21.3\% & \bfseries {\cellcolor{lightgray}} 18.5\% \\
 & \multirow[c]{2}{*}{\texttt{qf}} & 5.8\% & \bfseries 3.9\% & - & 13.1\% & \bfseries 10.8\% & - & \bfseries 36.4\% & 37.4\% & - \\
 &  & {\cellcolor{lightgray}} 10.7\% & \bfseries {\cellcolor{lightgray}} 8.5\% & {\cellcolor{lightgray}} - & {\cellcolor{lightgray}} 14.9\% & \bfseries {\cellcolor{lightgray}} 12.0\% & {\cellcolor{lightgray}} - & \bfseries {\cellcolor{lightgray}} 36.1\% & {\cellcolor{lightgray}} 36.4\% & {\cellcolor{lightgray}} - \\
 & \multirow[c]{2}{*}{\texttt{qt}} & 5.9\% & \bfseries 4.0\% & - & 12.6\% & \bfseries 10.5\% & - & \bfseries 36.6\% & 37.7\% & - \\
 &  & {\cellcolor{lightgray}} 11.0\% & \bfseries {\cellcolor{lightgray}} 8.8\% & {\cellcolor{lightgray}} - & {\cellcolor{lightgray}} 14.5\% & \bfseries {\cellcolor{lightgray}} 11.8\% & {\cellcolor{lightgray}} - & \bfseries {\cellcolor{lightgray}} 36.4\% & {\cellcolor{lightgray}} 36.8\% & {\cellcolor{lightgray}} - \\
\multirow[c]{8}{*}{Trasf.} & \multirow[c]{2}{*}{\texttt{pf}} & 0.8\% & \bfseries 0.4\% & 5.0\% & 1.6\% & \bfseries 1.2\% & 7.2\% & \bfseries 3.5\% & 3.7\% & 12.8\% \\
 &  & {\cellcolor{lightgray}} 2.9\% & \bfseries {\cellcolor{lightgray}} 1.1\% & {\cellcolor{lightgray}} 5.3\% & {\cellcolor{lightgray}} 1.4\% & \bfseries {\cellcolor{lightgray}} 1.3\% & {\cellcolor{lightgray}} 7.1\% & \bfseries {\cellcolor{lightgray}} 3.9\% & {\cellcolor{lightgray}} 4.2\% & {\cellcolor{lightgray}} 12.8\% \\
 & \multirow[c]{2}{*}{\texttt{pt}} & 0.8\% & \bfseries 0.4\% & 5.0\% & 1.6\% & \bfseries 1.2\% & 7.2\% & \bfseries 3.5\% & 3.7\% & 12.8\% \\
 &  & {\cellcolor{lightgray}} 2.9\% & \bfseries {\cellcolor{lightgray}} 1.1\% & {\cellcolor{lightgray}} 5.3\% & {\cellcolor{lightgray}} 1.4\% & \bfseries {\cellcolor{lightgray}} 1.3\% & {\cellcolor{lightgray}} 7.1\% & \bfseries {\cellcolor{lightgray}} 3.9\% & {\cellcolor{lightgray}} 4.2\% & {\cellcolor{lightgray}} 12.8\% \\
 & \multirow[c]{2}{*}{\texttt{qf}} & 3.4\% & \bfseries 2.1\% & - & 10.4\% & \bfseries 8.5\% & - & \bfseries 20.7\% & 20.8\% & - \\
 &  & {\cellcolor{lightgray}} 7.2\% & \bfseries {\cellcolor{lightgray}} 5.5\% & {\cellcolor{lightgray}} - & {\cellcolor{lightgray}} 11.3\% & \bfseries {\cellcolor{lightgray}} 9.2\% & {\cellcolor{lightgray}} - & {\cellcolor{lightgray}} 20.9\% & \bfseries {\cellcolor{lightgray}} 20.5\% & {\cellcolor{lightgray}} - \\
 & \multirow[c]{2}{*}{\texttt{qt}} & 3.2\% & \bfseries 1.8\% & - & 9.5\% & \bfseries 8.1\% & - & \bfseries 21.4\% & 21.7\% & - \\
 &  & {\cellcolor{lightgray}} 6.8\% & \bfseries {\cellcolor{lightgray}} 4.6\% & {\cellcolor{lightgray}} - & {\cellcolor{lightgray}} 10.2\% & \bfseries {\cellcolor{lightgray}} 8.7\% & {\cellcolor{lightgray}} - & {\cellcolor{lightgray}} 21.8\% & \bfseries {\cellcolor{lightgray}} 21.5\% & {\cellcolor{lightgray}} - \\
\hline

\end{tabular}

}
\caption{Pre-power flow TRMAE of \textsc{DC-IPOPT} and  \textsc{CANOS} across different grid sizes. The non-shaded rows are \textsc{FullTop} results, while the shaded rows are \textsc{TopDrop} results. For each grid and each predicted value, we highlight the best performing model in bold.}
\label{tab:pre_pf_supervised_metrics}
\end{table*}

Table \ref{tab:pre_pf_supervised_metrics} compares the TRMAE of \textsc{DC-IPOPT} and  \textsc{CANOS} variants on the \textsc{FullTop} and \textsc{TopDrop} datasets across different grid sizes. Note that the \textsc{DC-IPOPT} solution before power flow does not provide reactive power components, so the corresponding TRMAE for the branches and generator is omitted.  This comparison contextualizes the errors introduced by \textit{learning} to solve AC-OPF problems against the backdrop of the well-documented errors from using the DC-OPF approximation. Four patterns of results are evident.
First, across almost all datasets and features, \textsc{CANOS} outperforms \textsc{DC-IPOPT} by a large margin. The only exception is the real power flows on the lines of the 10,000-bus grid where \textsc{DC-IPOPT} and \textsc{CANOS} attain comparable scores. Second, both \textsc{DC-IPOPT} and \textsc{CANOS} exhibit similar performance across \textsc{FullTop} and \textsc{TopDrop}, with the exception of the small 500-bus grid where \textsc{CANOS} does see degradation in performance. Third, both \textsc{DC-IPOPT} and \textsc{CANOS}'s supervised performance degrade with the size of the grid. Fourth, \textsc{Deep-CANOS} outperforms the \textsc{Wide-CANOS} across all features.\footnote{See Appendix \ref{apdx:additional_supervised_metrics} for MSE metrics.}

\subsubsection{Feasibility}
\begin{table*}[!t]
\centering
\resizebox{\textwidth}{!}{\begin{tabular}{lrr|rr|rr}
\hline
 & \multicolumn{2}{c|}{500} & \multicolumn{2}{c|}{2000} & \multicolumn{2}{c}{10000} \\
 & \textsc{Wide-CANOS} & \textsc{Deep-CANOS}$_{48}$ & \textsc{Deep-CANOS}$_{48}$ & \textsc{Deep-CANOS}$_{60}$ & \textsc{Deep-CANOS}$_{48}$ & \textsc{Deep-CANOS}$_{60}$ \\
\hline
\multirow[c]{2}{*}{Branch thermal limit from} & 2.15e-06 & \bfseries 4.39e-07 & \bfseries 2.60e-09 & 3.97e-09 & \bfseries 5.09e-07 & 5.99e-07 \\
 & {\cellcolor{lightgray}} 2.88e-05 & \bfseries {\cellcolor{lightgray}} 9.98e-06 & {\cellcolor{lightgray}} 5.97e-07 & \bfseries {\cellcolor{lightgray}} 3.59e-07 & {\cellcolor{lightgray}} 2.32e-06 & \bfseries {\cellcolor{lightgray}} 2.26e-06 \\
\multirow[c]{2}{*}{Branch thermal limit to} & 5.45e-07 & \bfseries 4.83e-08 & \bfseries 2.14e-07 & 2.52e-07 & \bfseries 2.24e-07 & 3.26e-07 \\
 & {\cellcolor{lightgray}} 2.74e-05 & \bfseries {\cellcolor{lightgray}} 5.65e-06 & {\cellcolor{lightgray}} 8.06e-07 & \bfseries {\cellcolor{lightgray}} 5.01e-07 & {\cellcolor{lightgray}} 1.95e-06 & \bfseries {\cellcolor{lightgray}} 1.90e-06 \\
Branch voltage angle difference & \bfseries 0.00e+00 & \bfseries 0.00e+00 & \bfseries 0.00e+00 & \bfseries 0.00e+00 & 3.05e-10 & \bfseries 0.00e+00 \\
\multirow[c]{2}{*}{Reactive power balance bus} & {\cellcolor{lightgray}} 2.21e-03 & \bfseries {\cellcolor{lightgray}} 1.82e-03 & {\cellcolor{lightgray}} 5.35e-03 & \bfseries {\cellcolor{lightgray}} 4.20e-03 & \bfseries {\cellcolor{lightgray}} 7.98e-03 & {\cellcolor{lightgray}} 8.33e-03 \\
 & 4.75e-03 & \bfseries 4.11e-03 & 5.26e-03 & \bfseries 4.76e-03 & \bfseries 7.82e-03 & 8.18e-03 \\
\multirow[c]{2}{*}{Real power balance bus} & {\cellcolor{lightgray}} 7.17e-03 & \bfseries {\cellcolor{lightgray}} 5.30e-03 & {\cellcolor{lightgray}} 1.08e-02 & \bfseries {\cellcolor{lightgray}} 9.11e-03 & \bfseries {\cellcolor{lightgray}} 2.40e-02 & {\cellcolor{lightgray}} 2.53e-02 \\
 & 1.07e-02 & \bfseries 1.03e-02 & 1.03e-02 & \bfseries 9.81e-03 & \bfseries 2.18e-02 & 2.48e-02 \\
\hline
\end{tabular}
}
\caption{Pre-power flow Feasibility of \textsc{CANOS} solutions across different grid sizes. The non-shaded rows are \textsc{FullTop} results, while the shaded rows are \textsc{TopDrop} results. For each grid and each predicted value, we highlight the best performing model architecture in bold.}
\label{table:pre_pf_feasibility}
\end{table*}

Table \ref{table:pre_pf_feasibility} shows the constraint degree violations for the \textsc{FullTop} and \textsc{TopDrop} datasets across the three grid sizes. The pattern of results is remarkably consistent across the grid sizes and datasets. \textsc{CANOS} satisfies many constraints near-perfectly (e.g. voltage angle differences $\approx 10^{-9}$) across all of the datasets, even for the largest grid size and the \textsc{TopDrop} dataset. For all datasets, there are very small violations on the branch thermal limits. The only substantive violations are in the real and reactive bus power balance constraints which are on the order of $10^{-2}$.  This level of constraint is certainly tolerable in applications where DC-OPF solutions are used directly (e.g. year-long network planning, stochastic or security-constrained OPF \cite{Baker2021,Mancarella2020}). Unlike the supervised metrics in the previous section, there is not a huge disparity between \textsc{CANOS} performance as grid size increases. This adds further credibility to \textsc{CANOS} as a replacement for DC solvers for AC-OPF problems in real-world applications. However, for applications with stricter requirements on AC power balance, there is the option for further post-processing with power flow. 

\subsection{After Post-Processing With Power Flow}
We use power flow in two ways. First, we follow \cite{Zamzam2020} to use this as a feasibility restoration procedure for the outputs of \textsc{CANOS} where the primary constraint violations stem from real and reactive bus power imbalance. This procedure is substantially faster than solving an OPF problem so it could be used as a post-processing step without a prohibitive increase in processing time. Second, we use it to post-process the outputs of \textsc{DC-IPOPT}, expanding the DC-OPF solution to a full AC solution and enabling a comprehensive comparison of feasibility between \textsc{DC-IPOPT} and \textsc{CANOS}.
If the solutions from \textsc{CANOS} have smaller constraint violations than the DC-OPF counterpart, this forms a compelling argument to replace applications of DC-OPF with \textsc{CANOS} or similar models.

One possibility when running power flow is that the procedure does not converge. In Table \ref{tab:pf_convergence}, we show the convergence rates. The convergence rates are generally high (\> 98\%) across both \textsc{CANOS} and \textsc{DC-IPOPT}. We see more variability across dataset conditions than across between \textsc{CANOS} and \textsc{DC-IPOPT}. For example, both solvers have a convergence rate of 98.6\% for the 500-bus \textsc{TopDrop} dataset and a 100\% for the 10,000-bus \textsc{TopDrop} condition. In the subsequent metrics we only report values where power flow converged. This means that post-power flow metrics could be computed on potentially different subsets of examples if convergence rate across solvers isn't uniformly 100\%.

\begin{table}[ht]
\centering
\resizebox{0.5\textwidth}{!}{\begin{tabular}{lr|r|r|r}
\hline
 & \textsc{Wide-CANOS} & \textsc{Deep-CANOS}$_{48}$ & \textsc{Deep-CANOS}$_{60}$ & \textsc{DC-IPOPT} \\
\hline
\multirow[c]{2}{*}{500} & 100.0\% & 100.0\% & - & 100.0\% \\
 & {\cellcolor{lightgray}} 98.6\% & {\cellcolor{lightgray}} 98.6\% & {\cellcolor{lightgray}} - & {\cellcolor{lightgray}} 98.6\% \\
\multirow[c]{2}{*}{2000} & - & 100.0\% & 100.0\% & 100.0\% \\
 & {\cellcolor{lightgray}} - & {\cellcolor{lightgray}} 100.0\% & {\cellcolor{lightgray}} 100.0\% & {\cellcolor{lightgray}} 100.0\% \\
\multirow[c]{2}{*}{10000} & - & 100.0\% & 100.0\% & 100.0\% \\
 & {\cellcolor{lightgray}} - & {\cellcolor{lightgray}} 100.0\% & {\cellcolor{lightgray}} 100.0\% & {\cellcolor{lightgray}} 100.0\% \\
\hline
\end{tabular}

}
\caption{Percentage of power flow convergence when initialized with \textsc{CANOS} or \textsc{DC-IPOPT} solutions for different grid sizes.}
\label{tab:pf_convergence}
\end{table}

\subsubsection{Solution Fidelity / Supervised Metrics}

\begin{table*}[!t]
\centering
\resizebox{\textwidth}{!}{\begin{tabular}{llrrr|rrr|rrr}
\hline
 &  & \multicolumn{3}{c|}{500} & \multicolumn{3}{c|}{2000} & \multicolumn{3}{c}{10000} \\
 &  & \textsc{Wide-CANOS} & \textsc{Deep-CANOS}$_{48}$ & \textsc{DC-IPOPT} & \textsc{Deep-CANOS}$_{48}$ & \textsc{Deep-CANOS}$_{60}$ & \textsc{DC-IPOPT} & \textsc{Deep-CANOS}$_{48}$ & \textsc{Deep-CANOS}$_{60}$ & \textsc{DC-IPOPT} \\
\hline
\multirow[c]{4}{*}{Bus } & \multirow[c]{2}{*}{\texttt{va}} & 2.7\% & \bfseries 1.4\% & 74.9\% & 7.0\% & \bfseries 6.5\% & 117.6\% & \bfseries 32.7\% & 33.3\% & 632.4\% \\
 &  & {\cellcolor{lightgray}} 7.3\% & \bfseries {\cellcolor{lightgray}} 4.0\% & {\cellcolor{lightgray}} 75.2\% & {\cellcolor{lightgray}} 8.5\% & \bfseries {\cellcolor{lightgray}} 7.9\% & {\cellcolor{lightgray}} 117.9\% & \bfseries {\cellcolor{lightgray}} 34.1\% & {\cellcolor{lightgray}} 34.3\% & {\cellcolor{lightgray}} 631.8\% \\
 & \multirow[c]{2}{*}{\texttt{vm}} & 0.0\% & \bfseries 0.0\% & 7.7\% & 0.2\% & \bfseries 0.1\% & 6.9\% & 0.2\% & \bfseries 0.2\% & 5.9\% \\
 &  & {\cellcolor{lightgray}} 0.0\% & \bfseries {\cellcolor{lightgray}} 0.0\% & {\cellcolor{lightgray}} 7.7\% & {\cellcolor{lightgray}} 0.2\% & \bfseries {\cellcolor{lightgray}} 0.1\% & {\cellcolor{lightgray}} 6.9\% & {\cellcolor{lightgray}} 0.2\% & \bfseries {\cellcolor{lightgray}} 0.2\% & {\cellcolor{lightgray}} 5.9\% \\
\multirow[c]{4}{*}{Gen} & \multirow[c]{2}{*}{\texttt{pg}} & 0.3\% & \bfseries 0.1\% & 1.6\% & 0.5\% & \bfseries 0.4\% & 3.2\% & \bfseries 0.4\% & 0.4\% & 2.2\% \\
 &  & {\cellcolor{lightgray}} 0.5\% & \bfseries {\cellcolor{lightgray}} 0.3\% & {\cellcolor{lightgray}} 1.6\% & {\cellcolor{lightgray}} 0.5\% & \bfseries {\cellcolor{lightgray}} 0.5\% & {\cellcolor{lightgray}} 3.2\% & \bfseries {\cellcolor{lightgray}} 0.3\% & {\cellcolor{lightgray}} 0.4\% & {\cellcolor{lightgray}} 2.2\% \\
 & \multirow[c]{2}{*}{\texttt{qg}} & 6.7\% & \bfseries 4.6\% & 192.3\% & 23.0\% & \bfseries 19.4\% & 236.4\% & 30.6\% & \bfseries 30.5\% & 203.8\% \\
 &  & {\cellcolor{lightgray}} 14.5\% & \bfseries {\cellcolor{lightgray}} 10.8\% & {\cellcolor{lightgray}} 191.7\% & {\cellcolor{lightgray}} 25.8\% & \bfseries {\cellcolor{lightgray}} 21.2\% & {\cellcolor{lightgray}} 235.8\% & \bfseries {\cellcolor{lightgray}} 30.7\% & {\cellcolor{lightgray}} 30.7\% & {\cellcolor{lightgray}} 204.0\% \\
\multirow[c]{8}{*}{Line } & \multirow[c]{2}{*}{\texttt{pf}} & 1.8\% & \bfseries 0.8\% & 193.5\% & 2.6\% & \bfseries 2.4\% & 158.3\% & \bfseries 2.0\% & 2.0\% & 136.1\% \\
 &  & {\cellcolor{lightgray}} 4.3\% & \bfseries {\cellcolor{lightgray}} 2.5\% & {\cellcolor{lightgray}} 193.1\% & {\cellcolor{lightgray}} 3.1\% & \bfseries {\cellcolor{lightgray}} 2.9\% & {\cellcolor{lightgray}} 157.7\% & {\cellcolor{lightgray}} 2.1\% & \bfseries {\cellcolor{lightgray}} 2.0\% & {\cellcolor{lightgray}} 136.1\% \\
 & \multirow[c]{2}{*}{\texttt{pt}} & 1.8\% & \bfseries 0.8\% & 23.8\% & 2.6\% & \bfseries 2.4\% & 29.9\% & \bfseries 2.0\% & 2.0\% & 33.8\% \\
 &  & {\cellcolor{lightgray}} 4.3\% & \bfseries {\cellcolor{lightgray}} 2.5\% & {\cellcolor{lightgray}} 23.6\% & {\cellcolor{lightgray}} 3.1\% & \bfseries {\cellcolor{lightgray}} 2.9\% & {\cellcolor{lightgray}} 30.2\% & {\cellcolor{lightgray}} 2.1\% & \bfseries {\cellcolor{lightgray}} 2.0\% & {\cellcolor{lightgray}} 33.8\% \\
 & \multirow[c]{2}{*}{\texttt{qf}} & 3.2\% & \bfseries 2.0\% & 23.9\% & 8.9\% & \bfseries 7.6\% & 30.0\% & 14.7\% & \bfseries 14.6\% & 33.8\% \\
 &  & {\cellcolor{lightgray}} 6.7\% & \bfseries {\cellcolor{lightgray}} 5.1\% & {\cellcolor{lightgray}} 23.6\% & {\cellcolor{lightgray}} 10.6\% & \bfseries {\cellcolor{lightgray}} 8.5\% & {\cellcolor{lightgray}} 30.2\% & {\cellcolor{lightgray}} 15.1\% & \bfseries {\cellcolor{lightgray}} 14.6\% & {\cellcolor{lightgray}} 33.8\% \\
 & \multirow[c]{2}{*}{\texttt{qt}} & 3.4\% & \bfseries 2.2\% & 175.0\% & 8.5\% & \bfseries 7.3\% & 169.6\% & 14.7\% & \bfseries 14.6\% & 133.8\% \\
 &  & {\cellcolor{lightgray}} 7.1\% & \bfseries {\cellcolor{lightgray}} 5.4\% & {\cellcolor{lightgray}} 175.0\% & {\cellcolor{lightgray}} 10.2\% & \bfseries {\cellcolor{lightgray}} 8.2\% & {\cellcolor{lightgray}} 169.1\% & {\cellcolor{lightgray}} 15.1\% & \bfseries {\cellcolor{lightgray}} 14.7\% & {\cellcolor{lightgray}} 133.7\% \\
\multirow[c]{8}{*}{Trasf.} & \multirow[c]{2}{*}{\texttt{pf}} & 0.7\% & \bfseries 0.3\% & 201.4\% & 0.7\% & \bfseries 0.7\% & 118.3\% & \bfseries 1.1\% & 1.1\% & 186.9\% \\
 &  & {\cellcolor{lightgray}} 1.9\% & \bfseries {\cellcolor{lightgray}} 0.9\% & {\cellcolor{lightgray}} 200.4\% & {\cellcolor{lightgray}} 0.8\% & \bfseries {\cellcolor{lightgray}} 0.8\% & {\cellcolor{lightgray}} 118.2\% & \bfseries {\cellcolor{lightgray}} 1.1\% & {\cellcolor{lightgray}} 1.1\% & {\cellcolor{lightgray}} 187.0\% \\
 & \multirow[c]{2}{*}{\texttt{pt}} & 0.7\% & \bfseries 0.3\% & 7.1\% & 0.7\% & \bfseries 0.7\% & 7.3\% & \bfseries 1.1\% & 1.1\% & 17.1\% \\
 &  & {\cellcolor{lightgray}} 1.9\% & \bfseries {\cellcolor{lightgray}} 0.9\% & {\cellcolor{lightgray}} 7.6\% & {\cellcolor{lightgray}} 0.8\% & \bfseries {\cellcolor{lightgray}} 0.8\% & {\cellcolor{lightgray}} 7.3\% & \bfseries {\cellcolor{lightgray}} 1.1\% & {\cellcolor{lightgray}} 1.1\% & {\cellcolor{lightgray}} 17.1\% \\
 & \multirow[c]{2}{*}{\texttt{qf}} & 3.2\% & \bfseries 1.9\% & 7.1\% & 7.8\% & \bfseries 6.4\% & 7.3\% & 17.5\% & 17.2\% & \bfseries 17.1\% \\
 &  & {\cellcolor{lightgray}} 6.3\% & \bfseries {\cellcolor{lightgray}} 4.8\% & {\cellcolor{lightgray}} 7.6\% & {\cellcolor{lightgray}} 9.0\% & {\cellcolor{lightgray}} 7.4\% & \bfseries {\cellcolor{lightgray}} 7.3\% & {\cellcolor{lightgray}} 17.8\% & {\cellcolor{lightgray}} 17.2\% & \bfseries {\cellcolor{lightgray}} 17.1\% \\
 & \multirow[c]{2}{*}{\texttt{qt}} & 2.8\% & \bfseries 1.5\% & 231.8\% & 7.3\% & \bfseries 6.2\% & 119.1\% & 18.3\% & \bfseries 18.0\% & 182.6\% \\
 &  & {\cellcolor{lightgray}} 6.0\% & \bfseries {\cellcolor{lightgray}} 4.0\% & {\cellcolor{lightgray}} 233.5\% & {\cellcolor{lightgray}} 8.3\% & \bfseries {\cellcolor{lightgray}} 7.0\% & {\cellcolor{lightgray}} 119.5\% & {\cellcolor{lightgray}} 18.6\% & \bfseries {\cellcolor{lightgray}} 18.1\% & {\cellcolor{lightgray}} 182.8\% \\
\hline
\end{tabular}
}
\caption{Post-power flow TRMAE of \textsc{DC-IPOPT} and  \textsc{CANOS} across different grid sizes. The non-shaded rows are \textsc{FullTop} results, while the shaded rows are \textsc{TopDrop} results. For each grid and each predicted value, we highlight the best performing model in bold.}
\label{tab:post_pf_supervised_metrics}
\end{table*}

As before, we can compute supervised metrics comparing the distance between \textsc{CANOS} and \textsc{DC-IPOPT} solutions against the \textsc{AC-IPOPT} solutions. Unlike the results before post-processing with power flow, the outputs of \textsc{DC-IPOPT} can now be evaluated against all solution features.

Table~\ref{tab:post_pf_supervised_metrics} shows a continued advantage of \textsc{CANOS} over \textsc{DC-IPOPT} after solving the power flow equations for both the \textsc{FullTop} and \textsc{TopDrop} variants.  This holds across all features of the solution. We also continue to observe better performance of \textsc{Deep-CANOS} over \textsc{Wide-CANOS} and generally higher errors on \textsc{TopDrop} relative to \textsc{FullTop}. The most striking result, however, is that several of the \textsc{DC-IPOPT} metrics are above 100\% (even reaching as high as 600\%) after power flow. This underscores the poor approximating quality of using DC-OPF solvers \cite{yang2017solving}: before power flow the solutions are incomplete; after power flow they are highly inaccurate and (as shown in the next section) infeasible.

\subsubsection{Feasibility}

\begin{table*}[!t]
\centering
\resizebox{\textwidth}{!}{\begin{tabular}{lrrr|rrr|rrr}
\hline
 & \multicolumn{3}{c|}{500} & \multicolumn{3}{c|}{2000} & \multicolumn{3}{c}{10000} \\
 & \textsc{Wide-CANOS} & \textsc{Deep-CANOS}$_{48}$ & DC OPF & \textsc{Deep-CANOS}$_{48}$ & \textsc{Deep-CANOS}$_{60}$ & DC OPF & \textsc{Deep-CANOS}$_{48}$ & \textsc{Deep-CANOS}$_{60}$ & DC OPF \\
\hline
\multirow[c]{2}{*}{Branch thermal limit from} & 7.54e-06 & \bfseries 5.24e-06 & 2.25e-02 & 3.11e-05 & \bfseries 2.65e-05 & 1.10e-01 & 1.36e-05 & \bfseries 1.35e-05 & 6.54e-02 \\
 & {\cellcolor{lightgray}} 1.14e-04 & \bfseries {\cellcolor{lightgray}} 6.41e-05 & {\cellcolor{lightgray}} 2.26e-02 & {\cellcolor{lightgray}} 3.59e-04 & \bfseries {\cellcolor{lightgray}} 3.29e-04 & {\cellcolor{lightgray}} 1.09e-01 & {\cellcolor{lightgray}} 2.69e-05 & \bfseries {\cellcolor{lightgray}} 2.64e-05 & {\cellcolor{lightgray}} 6.55e-02 \\
\multirow[c]{2}{*}{Branch thermal limit to} & 1.17e-06 & \bfseries 3.17e-07 & 3.59e-02 & 4.59e-05 & \bfseries 3.96e-05 & 1.17e-01 & 8.42e-06 & \bfseries 7.72e-06 & 6.65e-02 \\
 & {\cellcolor{lightgray}} 1.13e-04 & \bfseries {\cellcolor{lightgray}} 6.92e-05 & {\cellcolor{lightgray}} 3.60e-02 & {\cellcolor{lightgray}} 4.16e-04 & \bfseries {\cellcolor{lightgray}} 3.79e-04 & {\cellcolor{lightgray}} 1.17e-01 & {\cellcolor{lightgray}} 2.08e-05 & \bfseries {\cellcolor{lightgray}} 2.08e-05 & {\cellcolor{lightgray}} 6.65e-02 \\
\multirow[c]{2}{*}{Branch voltage angle difference} & \bfseries 0.00e+00 & \bfseries 0.00e+00 & \bfseries 0.00e+00 & \bfseries 0.00e+00 & \bfseries 0.00e+00 & \bfseries 0.00e+00 & \bfseries 0.00e+00 & \bfseries 0.00e+00 & \bfseries 0.00e+00 \\
 & \bfseries {\cellcolor{lightgray}} 0.00e+00 & \bfseries {\cellcolor{lightgray}} 0.00e+00 & \bfseries {\cellcolor{lightgray}} 0.00e+00 & {\cellcolor{lightgray}} 4.67e-09 & {\cellcolor{lightgray}} 5.07e-09 & \bfseries {\cellcolor{lightgray}} 2.63e-09 & \bfseries {\cellcolor{lightgray}} 0.00e+00 & \bfseries {\cellcolor{lightgray}} 0.00e+00 & \bfseries {\cellcolor{lightgray}} 0.00e+00 \\
\multirow[c]{2}{*}{Bus voltage bounds pq} & 1.62e-06 & 4.57e-07 & \bfseries 0.00e+00 & 5.55e-06 & 2.45e-06 & \bfseries 0.00e+00 & 2.03e-06 & \bfseries 1.92e-06 & 8.89e-05 \\
 & {\cellcolor{lightgray}} 3.73e-06 & {\cellcolor{lightgray}} 2.57e-06 & \bfseries {\cellcolor{lightgray}} 2.00e-06 & {\cellcolor{lightgray}} 8.16e-06 & {\cellcolor{lightgray}} 3.56e-06 & \bfseries {\cellcolor{lightgray}} 1.72e-07 & {\cellcolor{lightgray}} 2.24e-06 & \bfseries {\cellcolor{lightgray}} 2.17e-06 & {\cellcolor{lightgray}} 8.87e-05 \\
\multirow[c]{2}{*}{Bus voltage bounds pv} & \bfseries 0.00e+00 & \bfseries 0.00e+00 & \bfseries 0.00e+00 & 1.87e-06 & 2.39e-07 & \bfseries 0.00e+00 & 5.50e-07 & 5.49e-07 & \bfseries 6.31e-10 \\
 & \bfseries {\cellcolor{lightgray}} 3.17e-08 & {\cellcolor{lightgray}} 4.56e-08 & {\cellcolor{lightgray}} 2.92e-07 & {\cellcolor{lightgray}} 3.77e-06 & {\cellcolor{lightgray}} 7.48e-07 & \bfseries {\cellcolor{lightgray}} 1.09e-07 & {\cellcolor{lightgray}} 6.74e-07 & {\cellcolor{lightgray}} 6.46e-07 & \bfseries {\cellcolor{lightgray}} 8.05e-08 \\
\multirow[c]{2}{*}{Generator reactive power bounds slack} & \bfseries 0.00e+00 & \bfseries 0.00e+00 & \bfseries 0.00e+00 & \bfseries 0.00e+00 & \bfseries 0.00e+00 & \bfseries 0.00e+00 & 8.49e-02 & 9.40e-02 & \bfseries 0.00e+00 \\
 & \bfseries {\cellcolor{lightgray}} 0.00e+00 & \bfseries {\cellcolor{lightgray}} 0.00e+00 & {\cellcolor{lightgray}} 1.96e-04 & {\cellcolor{lightgray}} 3.71e-04 & {\cellcolor{lightgray}} 3.02e-04 & \bfseries {\cellcolor{lightgray}} 1.69e-04 & {\cellcolor{lightgray}} 9.41e-02 & {\cellcolor{lightgray}} 9.66e-02 & \bfseries {\cellcolor{lightgray}} 5.69e-05 \\
\multirow[c]{2}{*}{Generator real power bounds slack} & \bfseries 7.55e-02 & 8.28e-02 & 4.03 & \bfseries 4.98e-01 & 5.08e-01 & 10.8 & 4.51e-01 & \bfseries 4.20e-01 & 20.7 \\
 & {\cellcolor{lightgray}} 1.87e-01 & \bfseries {\cellcolor{lightgray}} 1.54e-01 & {\cellcolor{lightgray}} 4.03 & {\cellcolor{lightgray}} 6.19e-01 & \bfseries {\cellcolor{lightgray}} 6.08e-01 & {\cellcolor{lightgray}} 10.8 & \bfseries {\cellcolor{lightgray}} 4.32e-01 & {\cellcolor{lightgray}} 4.36e-01 & {\cellcolor{lightgray}} 20.7 \\
\hline

\end{tabular}
}
\caption{Post-power flow feasiblity comparison between \textsc{DC-IPOPT} and  \textsc{CANOS} across different grid sizes. The non-shaded rows are \textsc{FullTop} results, while the shaded rows are \textsc{TopDrop} results. For each grid and each post-power flow value, we highlight the best performing model in bold.}
\label{tab:post_pf_feasibility}
\end{table*}

Table~\ref{tab:post_pf_feasibility} shows the violable constraints after power flow. For many of the branch constraints, \textsc{DC-IPOPT} and \textsc{CANOS} produce no violations or minimal ones on the order of $10^{-6}$.  We see larger violations for the branch thermal limits by \textsc{DC-IPOPT} e.g. for the 500-bus \textsc{TopDrop} $0.0226$ (from direction) and $0.0360$ (to direction) compared to CANOS $6.41 \times 10^{-5}$ and $6.92 \times 10^{-5}$. By far, however, the largest violation is in the slack generator's real power bounds for both \textsc{CANOS} and \textsc{DC-IPOPT}. Although the best \textsc{CANOS} model produces high violations ($0.154$) compared to the thresholds used in \textsc{AC-IPOPT}, \textsc{DC-IPOPT} produces violations \textit{$25\times$} larger than that for the 500-bus grid: ($4.03$). For context on the magnitude of these violations, in real units this translates into a difference between 15 megawatts for \textsc{CANOS} versus 403 megawatts for \textsc{DC-IPOPT} on the slack generator which has a maximum output of $1164$ megawatts. These discrepancies are further exacerbated on the 2,000-bus grid where the \textsc{DC-IPOPT} real slack violation is $10.8$, but \textsc{CANOS} violations are only $0.608$ and the 10000-bus grid with $20.7$ and $0.432$ respectively.  This builds further evidence that \textsc{CANOS} is a more accurate and feasible alternative to \textsc{DC-IPOPT} in situations where approximate solvers are appropriate.

\subsection{Optimality}

\begin{table}[ht]
\centering
\resizebox{0.5\textwidth}{!}{\begin{tabular}{lr|r|r|r}
\hline
 & \textsc{Wide-CANOS} & \textsc{Deep-CANOS}$_{48}$ & \textsc{Deep-CANOS}$_{60}$ & \textsc{DC-IPOPT} \\
\hline
\multirow[c]{2}{*}{500} & 100.10\% & 99.98\% & - & 96.84\% \\
 & {\cellcolor{lightgray}} 100.21\% & {\cellcolor{lightgray}} 100.01\% & {\cellcolor{lightgray}} - & {\cellcolor{lightgray}} 96.82\% \\
\multirow[c]{2}{*}{2000} & - & 99.98\% & 99.95\% & 96.87\% \\
 & {\cellcolor{lightgray}} - & {\cellcolor{lightgray}} 99.96\% & {\cellcolor{lightgray}} 99.94\% & {\cellcolor{lightgray}} 96.87\% \\
\multirow[c]{2}{*}{10000} & - & 100.04\% & 100.04\% & 99.41\% \\
 & {\cellcolor{lightgray}} - & {\cellcolor{lightgray}} 100.04\% & {\cellcolor{lightgray}} 100.05\% & {\cellcolor{lightgray}} 99.41\% \\
\hline
\end{tabular}}
\caption{Ratio (in percentage) of \textsc{CANOS} and \textsc{DC-IPOPT} solutions' cost with respect to \textsc{AC-IPOPT} cost, before power flow. The non-shaded rows are \textsc{FullTop} results, while the shaded rows are \textsc{TopDrop} results.}
\label{tab:optimality_pre}
\end{table}

\begin{table}[ht]
\centering
\resizebox{0.5\textwidth}{!}{\begin{tabular}{lr|r|r|r}
\hline
 & \textsc{Wide-CANOS} & \textsc{Deep-CANOS}$_{48}$ & \textsc{Deep-CANOS}$_{60}$ & \textsc{DC-IPOPT} \\
\hline
\multirow[c]{2}{*}{500} & 100.08\% & 99.99\% & - & 97.95\% \\
 & {\cellcolor{lightgray}} 100.16\% & {\cellcolor{lightgray}} 100.01\% & {\cellcolor{lightgray}} - & {\cellcolor{lightgray}} 97.93\% \\
\multirow[c]{2}{*}{2000} & - & 99.99\% & 99.97\% & 98.40\% \\
 & {\cellcolor{lightgray}} - & {\cellcolor{lightgray}} 99.98\% & {\cellcolor{lightgray}} 99.97\% & {\cellcolor{lightgray}} 98.40\% \\
\multirow[c]{2}{*}{10000} & - & 100.04\% & 100.04\% & 102.30\% \\
 & {\cellcolor{lightgray}} - & {\cellcolor{lightgray}} 100.03\% & {\cellcolor{lightgray}} 100.04\% & {\cellcolor{lightgray}} 102.30\% \\
\hline

\end{tabular}
}
\caption{Ratio (in percentage) of \textsc{CANOS} and \textsc{DC-IPOPT} solutions' cost with respect to \textsc{AC-IPOPT} cost, after power flow. The non-shaded rows are \textsc{FullTop} results, while the shaded rows are \textsc{TopDrop} results.}
\label{tab:optimality_post}
\end{table}

Before power flow, table \ref{tab:optimality_pre}, and after power flow, table \ref{tab:optimality_post}, show that the OPF solutions provided by \textsc{CANOS} are accurate to within 1\% of the reference objective cost. For \textsc{Deep-CANOS}, this accuracy is within 0.1\%. Furthermore, both \textsc{CANOS} variants are more accurate than \textsc{DC-IPOPT} which is only accurate within 3-4\%.

\subsection{Speed}

\begin{figure}
%\centering
\begin{subfigure}{1.\textwidth}
    \resizebox{0.5\textwidth}{!}{\includegraphics{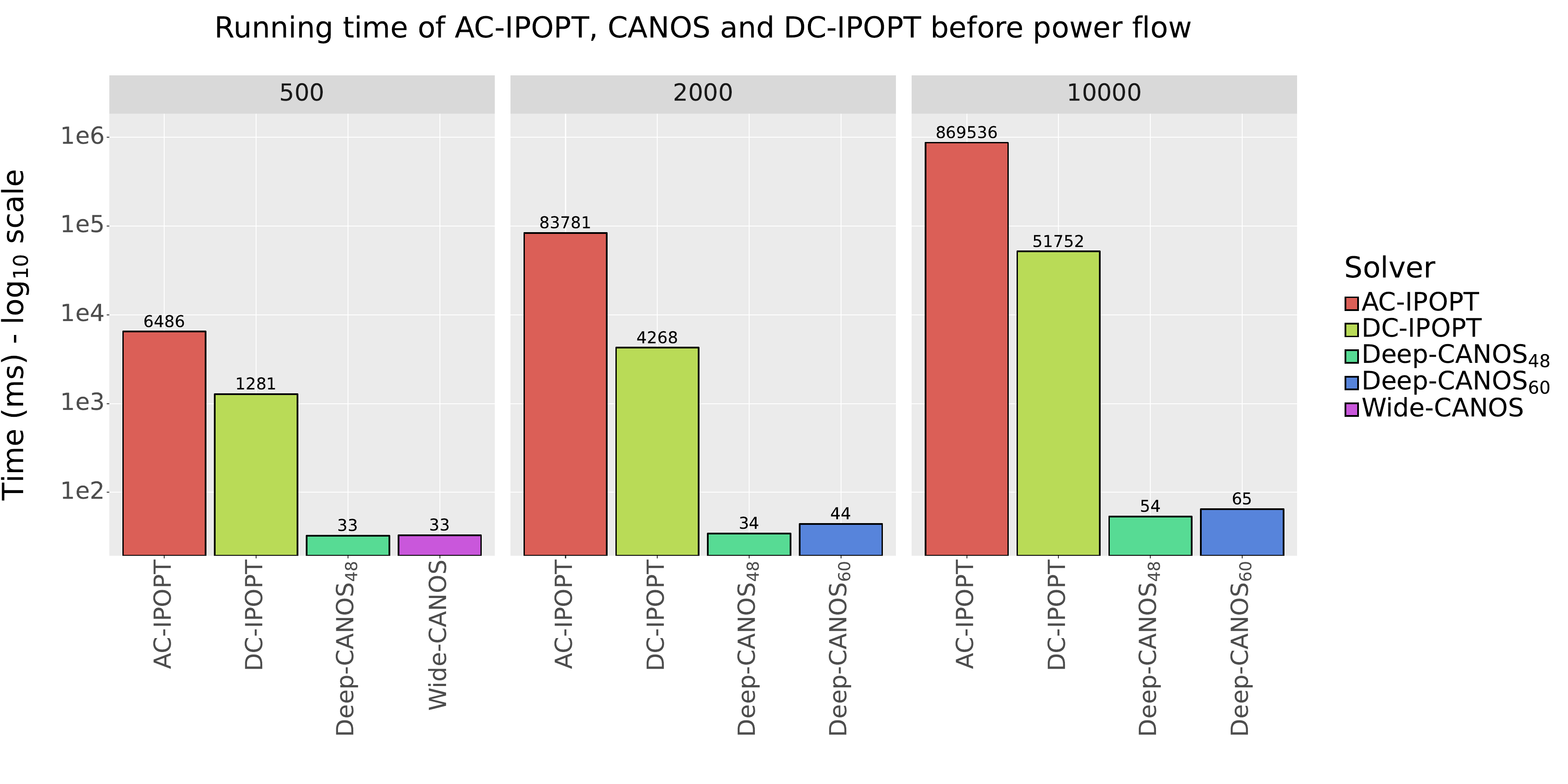}}
    \label{fig:pre_pf_speed}
\end{subfigure}
\begin{subfigure}{1.\textwidth}
    \resizebox{0.5\textwidth}{!}{\includegraphics{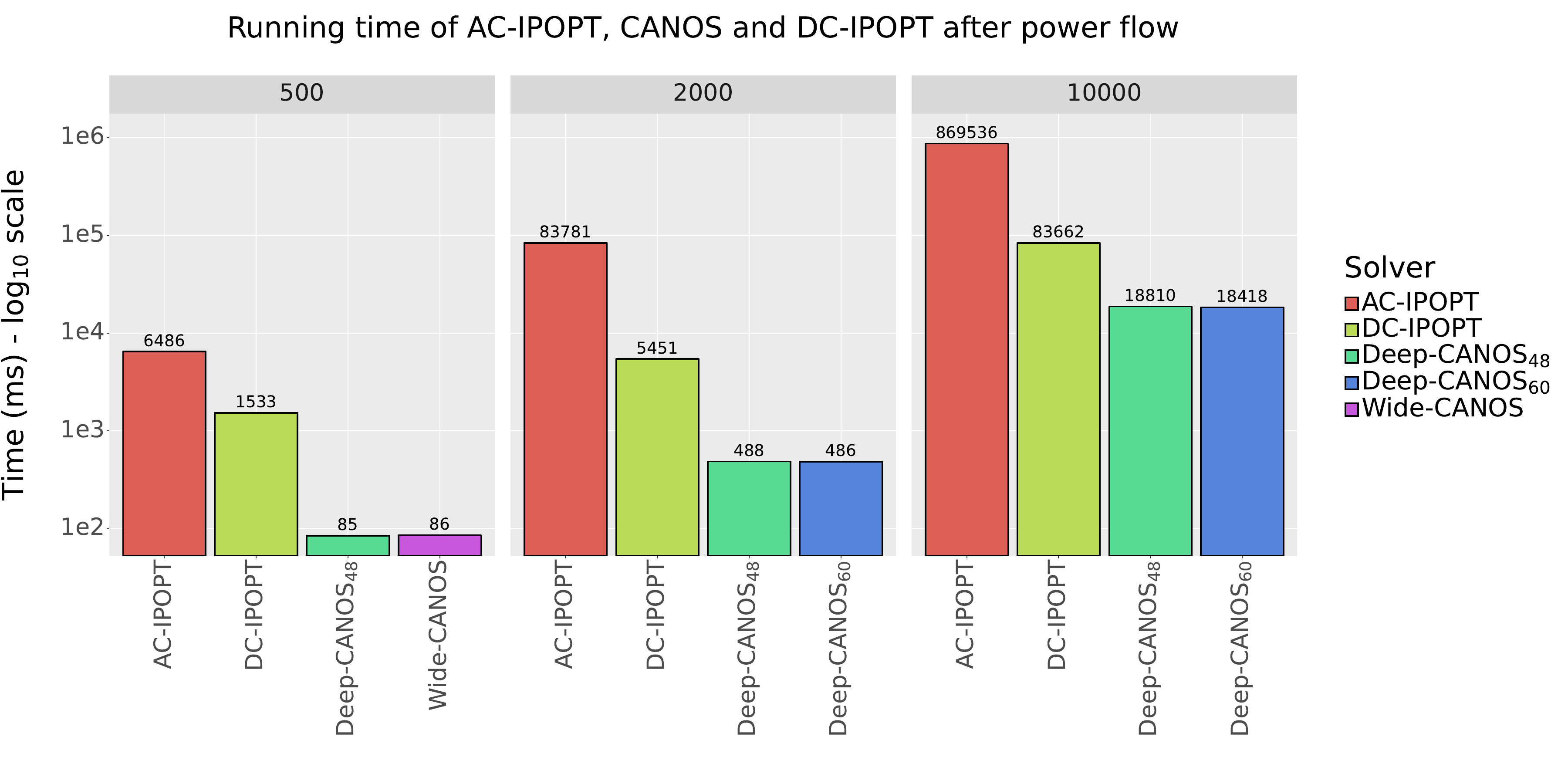}}
    \label{fig:post_pf_speed}
\end{subfigure}
\caption{Speed comparison of \textsc{AC-IPOPT}, \textsc{DC-IPOPT} and \textsc{CANOS}.}
\label{fig:speed}
\end{figure}

%\begin{figure}[!htb]
%    \centering
%    \resizebox{0.5\textwidth}{!}{\includegraphics{figures/speed_pre_pf}}
%    \caption{Speed comparison of \textsc{AC-IPOPT}, \textsc{DC-IPOPT} and \textsc{CANOS} before power flow.}
%    \label{fig:pre_pf_speed}
%\end{figure}

Figure \ref{fig:speed} (top) compares the processing times of \textsc{CANOS} against \textsc{DC-IPOPT} before either method performs post-processing. We focus on the results for the \textsc{TopDrop} datasets since speed does not significantly vary between dataset types.  For all grids, \textsc{CANOS} is faster than \textsc{DC-IPOPT}.  \textsc{CANOS} also scales more gracefully with grid size.  \textsc{DC-IPOPT} takes 1.3 seconds on the 500-bus grid, but 50 seconds on the 10,000-bus grid - almost a 50x increase. For \textsc{CANOS} with 48-message passing steps, solving a 500-bus grid takes roughly 33 ms and solving a 10,000-bus grid takes only 54 ms, demonstrating a sub-linear scaling with grid size. We also note that the differences between \textsc{CANOS} variants are fairly minute as well. These results show that not only is \textsc{CANOS} faster than \textsc{DC-IPOPT}, but also the speed-up is amplified on real-world grid sizes like the 10,000-bus grid. 

Figure \ref{fig:speed} (bottom) shows results for the total running times including the power flow post-processing. Here we can see the appeal of taking approximate methods and trying to make them AC-feasible instead of solving the full AC-OPF. Even with a relatively costly power flow step, both \textsc{CANOS} and \textsc{DC-IPOPT} run in a fraction of the time taken by \textsc{AC-IPOPT}.
\section{Conclusion}
\label{sec:conclusion}

Our experiments show that \textsc{CANOS} is more accurate and more feasible than using \textsc{DC-IPOPT} solver in AC-OPF problems. We've established this across a variety of grid sizes including a large 10,000-bus grid in the \textsc{TopDrop} setting which includes N-1 topological perturbations. We've established that \textsc{CANOS} is fast and scales well with grid size, running in as little as 65 ms for the 10,000-bus grid. Below we discuss limitations of the approach and future directions.

\subsection{Limitations of Machine Learning Approaches}
There are clear limitations to our approach that are worth addressing.  The most striking one is the inability to guarantee full AC-feasibility. This is a challenge for all ML-for-OPF methods, since enforcing complex constraints directly in the model architecture is non-trivial. Importantly, this is also a known limitation of DC-OPF solvers, arising from the formulation approximations which don't reflect the real equations.

One limitation that is specific to ML methods is the issue of data drift or distribution shift \cite{quinonero2008}. A model trained under one set of conditions may fail to generalize to future grid conditions and produce sub-optimal, inaccurate, or infeasible solutions.  This is not a fundamental blocker to the adoption of this technology for industrial applications, but it requires further work in two areas. First, more research is needed to empirically investigate how problematic data drift is in practice and how to alleviate it with fine-tuning or re-training. This could be conducted in both controlled synthetic scenarios and real-world conditions. Second, frameworks are needed to integrate regular monitoring to track these quantities and take corrective action as appropriate (either deferring to traditional solvers for a speed penalty or re-train/fine-tune models). In this respect, one promising property of \textsc{CANOS} (because it is robust to topological perturbations) is that it can be fine-tuned even if the grid was expanded to contain new units or branches.

\subsection{Future Directions}

Although our current results push the state-of-the-art for AC-OPF with topological variations, there are clear avenues for improvement. We used a now-standard approach (e.g. \cite{Fioretto2020,Zamzam2020}) of generating synthetic data by varying load perturbations independently . This load perturbation could be more effective by using the \textsc{OPFLearn} \cite{joswig2022opf} methodology or overlaying real-world patterns of demand onto the scenarios. The current approach would also benefit from a broader set of perturbations beyond just the demand for power, e.g. generator capacities or thermal ratings, to capture a wider set of scenarios. We could also keep the same types of perturbations and make them more extreme - such as dropping out more entities besides the N-1 considered here. These new scenarios could be used as further training data or as unseen evaluation examples to assess generalization capabilities.

Furthermore, recent years have seen a multitude of ML approaches brought to bear on learning OPF (see \cite{Khaloie2024}). Some of these approaches are directly competing with \textsc{CANOS}: they target AC-OPF with topological perturbations and make different architecture choices (e.g. \cite{Falconer2023} uses a Graph Convolutional Network with 3-4 layers instead of our larger Encode-Process-Decode GNN with 36-60 message-passing steps). Although our present focus was on comparing against traditional solvers, future work could compare these ML methods directly. Understanding the trade-offs between various architectures and hyperparameters' choices remains an important task\footnote{We believe \textsc{CANOS}'s strength lies in its size. Appendix \ref{apdx:model_size_analysis} shows \textsc{CANOS} performance improves greatly as we increase the number of message passing steps}.

Other approaches are independent of the neural architecture used and could be added as an extra component on top of \textsc{CANOS}. For example, we take a simple approach to penalizing constraint violations during training. Our results could be improved through more sophisticated approaches such as updating the constraint penalties using the Lagrangian approach \cite{Fioretto2020}.  Another promising route would be to incorporate the zero-gradient approach from \cite{Pan2021} to \textsc{CANOS} to take gradients through the post-processing power flow step and likely improve constraint satisfaction.

Beyond those bolt-on extensions to \textsc{CANOS} are larger opportunities in power grid optimization. AC-OPF is a root node problem in power grid optimization -- many branches stem from AC-OPF, either wrapping or extending it to form a more complex, computationally-demanding problem.  \textsc{CANOS} is a fast, accurate, near-optimal, near-feasible AC-OPF solver that's robust to topology perturbations. \textsc{CANOS}'s speed and accuracy make it a strong candidate to replace DC-OPF in problems such as stochastic OPF or sequential OPF where the sheer number of problems to be solved prevents the use of a full AC-OPF solver. Furthermore, because \textsc{CANOS} is robust to topological perturbations, it unlocks branches of power grid optimization that modify the grid connectivity, the most obvious case being SC-OPF, from which we adopt the N-1 \textsc{TopDrop} perturbations. It unlocks the potential to learn from multi-year, real-world grid data, which often includes altered or expanded topology. Furthermore, it unlocks mixed-integer AC-OPF problems like unit commitment or optimal transmission switching, which mix the continuous variables of standard AC-OPF with discrete variables that change the underlying grid connectivity. A particularly challenging application would be transmission network expansion planning, which combines the need for topology perturbations \textit{and} evaluating many scenarios. 

Although there will certainly be further improvements needed to reach these new applications, \textsc{CANOS} provides an extensible foundation to build from.

\bibliographystyle{ieeetr}
\bibliography{bib}

\vfill

\clearpage

\clearpage
\appendices
\label{sec:apdx}
\renewcommand{\theequation}{A-\arabic{equation}}
% redefine the command that creates the equation no.
\setcounter{equation}{0}  % reset counter 

\section{Data}
\label{apdx:data}
\subsection{Data Format}
Each example in our datasets contains a \texttt{grid}, representing the grid operating state (the input to the OPF problem) and a \texttt{solution}, containing the dispatch decisions for the grid (the OPF output). The data is structured as follows:
\vspace{10px}\\
\textbf{Grid nodes} \vspace*{4px}\\
\begin{tabular}{l l}
\textbf{Bus} &  \\
\texttt{base\_kv} & Base voltage (kV) \\
\texttt{bus\_type} & PQ, PV, reference or inactive \\
\texttt{vmin} ($v^l_i$) & Min voltage magnitude \\
\texttt{vmax} ($v^u_i$) & Max voltage magnitude \\
\textbf{Generator} &  \\
\texttt{mbase} & Total MVA base \\
\texttt{pg} & Initial real power generation as \\
& given in the pglib case \\
\texttt{pmin} ($\Re(S^{gl}_k)$) & Min real power generation \\
\texttt{pmax} ($\Re(S^{gu}_k)$) & Max real power generation \\
\texttt{qg} & Initial reactive power generation \\
& as given in the pglib case \\
\texttt{qmin} ($\Im(S^{gl}_k)$) & Min reactive power generation \\
\texttt{qmax} ($\Im(S^{gu}_k)$) & Max reactive power generation \\
\texttt{vg} & Initial voltage magnitude as \\
& given in the pglib case \\
\texttt{cost\_squared} ($c_{2k}$) & Coefficient of \texttt{pg}$^2$ in cost term \\
\texttt{cost\_linear} ($c_{1k}$) & Coefficient of \texttt{pg} in cost term \\
\texttt{cost\_offset} ($c_{0k}$) & Constant coefficient in cost term \\
\textbf{Load} &  \\
\texttt{pd} ($\Re(S^d_k)$) & Real power demand \\
\texttt{qd} ($\Im(S^d_k)$) & Reactive power demand \\
\textbf{Shunt} &  \\
\texttt{bs} ($\Im(Y^s_{k})$) & Shunt susceptance \\
\texttt{gs} ($\Re(Y^s_{k})$) & Shunt conductance \\
\end{tabular}\\
\vspace{10px}\\
\textbf{Grid edges} \vspace*{4px}\\
\begin{tabular}{ll}
\textbf{AC-line} &  \\
\texttt{angmin} ($\theta^{\Delta l}_{ij}$) & Min angle difference between \\ &
from and to bus (radians) \\
\texttt{angmax} ($\theta^{\Delta u}_{ij}$) & Max angle difference between \\ &
from and to bus (radians) \\
\texttt{b\_fr} ($\Im(Y^c_{ij})$) & Line charging susceptance at \\
& from bus \\
\texttt{b\_to} ($\Im(Y^c_{ji})$) & Line charging susceptance at \\
& to bus \\
\texttt{br\_r} ($\Re(1 / Y_{ij})$) & Branch series resistance \\
\texttt{br\_x} ($\Im(1 / Y_{ij})$) & Branch series reactance \\
\texttt{rate\_a} ($s^u_{ij}$) & Long term thermal line rating \\
\texttt{rate\_b} & Short term thermal line rating \\
\texttt{rate\_c} & Emergency thermal line rating \\
\textbf{Transformer} &  \\
\multicolumn{2}{l}{All AC-line features, augmented with:}\\
\texttt{tap} ($|T_{ij}|$) & Branch off nominal turns ratio \\
\texttt{shift} ($\angle T_{ij}$) & Branch phase shift angle \\
\end{tabular}
\vspace{10px}\\
\textbf{Grid context} \vspace*{4px}\\
\begin{tabular}{ll}
\texttt{baseMVA} & The system wide MVA value for converting \\
& between mixed-units and p.u. unit values \\
\end{tabular}
\vspace{5px}\\
\textbf{Solution nodes} \vspace*{4px}\\
\begin{tabular}{l l}
\textbf{Bus} &  \\
\texttt{va} ($\angle (V_i V^*_j)$) & Voltage angle\\
\texttt{vm} ($|V_i|$) & Voltage magnitude\\
\textbf{Generator} &  \\
\texttt{pg} ($\Re(S^g_k)$) & Real power generation\\
\texttt{qg} ($\Im(S^g_k)$) & Reactive power generation\\
\end{tabular}
\vspace{10px}\\
\textbf{Solution edges} \vspace*{4px}\\
\begin{tabular}{l l}
\textbf{AC-line} &  \\
\texttt{pt} ($\Re(S_{ji})$) & Active power withdrawn at the to bus\\
\texttt{qt} ($\Im(S_{ji})$) & Reactive power withdrawn at the to bus\\
\texttt{pf} ($\Re(S_{ij})$) & Active power withdrawn at the from bus\\
\texttt{qf} ($\Im(S_{ij})$) & Reactive power withdrawn at the from bus\\
\textbf{Transformer} &  \\
\texttt{pt} ($\Re(S_{ji})$) & Active power withdrawn at the to bus\\
\texttt{qt} ($\Im(S_{ji})$) & Reactive power withdrawn at the to bus\\
\texttt{pf} ($\Re(S_{ij})$) & Active power withdrawn at the from bus\\
\texttt{qf} ($\Im(S_{ij})$) & Reactive power withdrawn at the from bus\\
\end{tabular}\\

\subsection{Grid Sizes}

We report the number of elements for each grid used in the paper. We refer to each grid by the number of buses in the grid, but use the original scenario names below. The reported values reflect the base case values from \textsc{pglib-OPF} before any topogical perturbations.

\begin{table}[h]
\resizebox{.5\textwidth}{!}{
\begin{tabular}{lllllll}
        %\hline
        Scenario & $|\mathcal{N}|$ & $|\mathcal{G}|$ & $|\mathcal{D}|$ & $|\mathcal{S}|$ & $|\mathcal{E}_l|$ & $|\mathcal{E}_t|$ \\
        \hline
        pglib\_opf\_case500\_goc &  500 & 171 & 281 & 31 & 536 & 192 \\
        pglib\_opf\_case2000\_goc &  2000 & 238 & 1010 & 124 & 2737 & 896 \\
        pglib\_opf\_case10000\_goc &  10000 & 2016 & 3984 & 510 & 10819 & 2374 \\
        \hline
    \end{tabular}}
\caption{Summary of the number of elements for each grid, with buses $|\mathcal{N}|$, generators $|\mathcal{G}|$, loads $|\mathcal{D}|$, shunts $|\mathcal{S}|$, and edges $|\mathcal{E}_l|$ (AC lines), $|\mathcal{E}_t|$ (transformers).}
\label{tab:grid_sizes}
\end{table}

\clearpage
\section{Power Flow} 
\label{apdx:power_flow}

In what follows we provide a brief overview of power flow. For a comprehensive description, see~\cite{mccalleyt7}. Note also that the traditional method described here and adopted by \texttt{pandapower} solver have known limitations~\cite{ZENG2023108905}.

Power flow (or load flow) is the process of finding bus voltage angles and magnitudes which are solutions of the bus power balance equation~\eqref{eq:constr4}, for all buses.
%%%

For the purpose of power flow, it is easier to rewrite it so that all quantities depending on the voltages are on the right hand side of the equation, and separate the real and reactive components:
\begin{align}
\underbrace{\sum_{k\in\G_i} \Re(S^g_k) - \sum_{k\in\L_i} \Re(S^d_k)}_{P_i} &= \sum_{(i,j) \in \E_i \cup \E_i^R } p_{ij} + \sum_{k\in\S_i} g_k^s | V_i|^2 \label{eq:p_balance}\\
\underbrace{\sum_{k\in\G_i} \Im(S^g_k) - \sum_{k\in\L_i} \Im(S^d_k)}_{Q_i} &= \sum_{(i,j) \in \E_i \cup \E_i^R } q_{ij} - \sum_{k\in\S_i} b_k^s| V_i|^2 \label{eq:q_balance}
\end{align}
where $p_{ij} =\Re(S_{ij})$, and $q_{ij} =\Im(S_{ij})$ are the branch flow complex coefficients, and $g_k^s = \Re(Y_k^s)$, and $b_k^s = \Im(Y_k^s)$ are shunts' conductance and susceptance, the complex coefficient of the shunt admittance.

At each bus, we have four variables: the total active power P, the total reactive power Q, the voltage angle $\theta$, and the voltage magnitude $|V|$. If two of them are known, the two remaining can be determined using \eqref{eq:p_balance}--\eqref{eq:q_balance}.

\begin{table}
    \centering
        \begin{tabular}{l|l|l}
        Bus type & Known Variables & Unknown variables \\
        \hline
        PV & $P$, $|V|$ & $Q$, $\theta$ \\
        PQ & $P$, $Q$ & $|V|$, $\theta$ \\
        Slack & $V$, $\theta$ & $P$, $Q$
        \end{tabular}
    \caption{Buses classification for power flow analysis.}
    \label{tab:bus_types}
\end{table}

Buses are classified according to the known/unknown variables, as shown by table \ref{tab:bus_types}. PV buses are voltage controlled buses, where at least a generator is connected. PQ buses are all load buses and buses with nothing attached. The Slack bus\footnote{Other approaches with distributed slack are possible, but we restrict the discussion to a single slack.} is a generator bus selected to compensate for the network losses. The generator bus with the larger power capacity is typically chosen for this role, and its voltage angle is set to 0, thus acting also as the reference bus.\footnote{
It must be understood that the slack bus is an artifact of the procedure. Before solving the equations, bus voltages and angles are unknown, hence the line losses (which are a function of the branch flows) can't be computed until the end of the procedure. The power of at least one bus must not be set, so that it can compensate for these line losses.}

Let $N_{PQ}$ be be number of PQ buses and $N_{PV}$ the number of PV buses.
Each PQ bus has two uknowns: $|V|$ and $\theta$. These can be derived using the two equations ~\eqref{eq:p_balance}--\eqref{eq:q_balance} since at these buses both P and Q are known.
Each PV bus has a single unknown, $\theta$. Since P is known, we can derive it using the real power balance equation ~\eqref{eq:p_balance}.
Therefore, with a system of $2N_{PQ} + N_{PV}$ equations, all bus voltages can be determined by means of a traditional root-finding method, such as Newthon-Raphson or Gauss-Siedel.\footnote{In recent years, Holographic Embedding Load Flow~\cite{6344759} has been proposed as a non-iterative, deterministic method to solve power flow.}

Once they are known, the total reactive power at each PV bus, Q, can be found using~\eqref{eq:q_balance}.
The bus total reactive power is then distributed among connected generators using different heuristics, typically proportionally to their capacity.

Likewise, the slack real and reactive power P, Q can be calculated via ~\eqref{eq:p_balance}--\eqref{eq:q_balance}.

During the procedure, generators' reactive bounds can be violated. Power flow solvers typically provide an option to enforce them by limiting generators outside range to their maximum capacity\footnote{see also eq.~5 - 6a-6e in \cite{Baker2021}.} and converting the corresponding bus to a PQ bus.
This is the version of power flow we use in our analysis.

Therefore, after power flow, the following holds:
\begin{itemize}
    \item All PQ buses have new voltage values, $\texttt{va'}$, $\texttt{vm'}$.
    \item All PV buses have new voltage angles $\texttt{va'}$. Note that if reactive limits have been enforced and the bus has been limited, $\texttt{vm'}$ can show changes as well.
    \item Generators have new reactive power $\texttt{qg'}$.
    \item Slack generator(s) have new real and reactive power $\texttt{pg'}$, $\texttt{qg'}$. 
    \item New values for the voltage lead to new values for the branch flows, $\texttt{pf'}$, $\texttt{qf'}$, $\texttt{pt'}$, $\texttt{qt'}$.
\end{itemize}

\textsc{CANOS} pre-power flow predictions $\texttt{va}$, $\texttt{vm}$, $\texttt{pg}$, $\texttt{qg}$ are all guaranteed to be within bounds~(\eqref{eq:constr2}--\eqref{eq:constr3}) by the sigmoid mapping in output layer~\eqref{eq:sigmoid_bound}. Likewise, the \textsc{DC-IPOPT} $\texttt{va}$, $\texttt{vm}$, $\texttt{pg}$ baseline solutions are guaranteed to satisfy bounding constraints by the constrained-optimization procedure (reactive power is ignored in the DC approximation).

The power flow algorithm that we apply only guarantees that reactive limits are satisfied for all but the slack generator(s). Bus voltage bounds~\eqref{eq:constr3} are not enforced and can potentially be violated. Likewise, slack generator(s) can violate active and reactive bounds. When such violations are observed, the power flow solution can be infeasible.

We now summarize the expectations on constraints violations (feasibility metrics):
\begin{itemize}
    \item Buses
        \begin{itemize}
            \item voltage bounds~\eqref{eq:constr3}: can be violated by all PQ buses, and by some PV buses as a result of enforcing reactive limits. Violations signal an infeasible solution. The slack bus cannot violate bounds since the voltage is fixed to pre-power flow values.
            \item bus power balance: guaranteed to be satisfied at numerical accuracy of \texttt{pandapower} solver.
        \end{itemize}
    \item Generators
        \begin{itemize}
            \item active power bounds~\eqref{eq:constr2}: non-slack generators have the power fixed to the pre-power flow values. Hence, no violation is expected. The slack generator(s) can instead violate power bounds, leading to infeasible solutions.
            \item reactive power bounds~\eqref{eq:constr2}: for non-slack generators, \texttt{enforce\_q\_limit} guarantees that they are satisfied. The slack generator(s) can instead violate power bounds, leading to infeasible solutions.
        \end{itemize}
    \item Branches
        \begin{itemize}
            \item thermal limits~\eqref{eq:constr7}: All branches can violate thermal limits, before ($\texttt{pf}$, $\texttt{qf}$, $\texttt{pt}$, $\texttt{qt}$) and after power flow ($\texttt{pf'}$, $\texttt{qf'}$, $\texttt{pt'}$, $\texttt{qt'}$)
            \item voltage angle difference~\eqref{eq:constr8}: All branches can potentially violate it, before and after power flow.
        \end{itemize}
\end{itemize}

Note also that the cost of the solution~\eqref{eq:cost} is a function of the real power only. Since only the slack power can change after power flow, the slack is fully responsible for differences between pre-power flow and post-power flow cost.

\clearpage
\onecolumn
\section{Additional Metrics}
\label{apdx:additional_supervised_metrics}

\subsection{MSE}
\begin{table}[!h]
\centering
\resizebox{\textwidth}{!}{\begin{tabular}{llrrr|rrr|rrr}
\hline
 &  & \multicolumn{3}{c|}{500} & \multicolumn{3}{c|}{2000} & \multicolumn{3}{c}{10000} \\
 &  & \textsc{Wide-CANOS} & \textsc{Deep-CANOS}$_{48}$ & \textsc{DC-IPOPT} & \textsc{Deep-CANOS}$_{48}$ & \textsc{Deep-CANOS}$_{60}$ & \textsc{DC-IPOPT} & \textsc{Deep-CANOS}$_{48}$ & \textsc{Deep-CANOS}$_{60}$ & \textsc{DC-IPOPT} \\
\hline
\multirow[c]{4}{*}{Bus } & \multirow[c]{2}{*}{\texttt{va}} & 3.05e-05 & \bfseries 1.50e-06 & 4.38e-03 & 3.65e-05 & \bfseries 3.22e-05 & 2.13e-02 & 1.49e-04 & \bfseries 1.48e-04 & 1.29e-02 \\
 &  & {\cellcolor{lightgray}} 1.59e-04 & \bfseries {\cellcolor{lightgray}} 4.70e-05 & {\cellcolor{lightgray}} 4.35e-03 & {\cellcolor{lightgray}} 1.16e-04 & \bfseries {\cellcolor{lightgray}} 1.06e-04 & {\cellcolor{lightgray}} 2.14e-02 & {\cellcolor{lightgray}} 1.61e-04 & \bfseries {\cellcolor{lightgray}} 1.60e-04 & {\cellcolor{lightgray}} 1.29e-02 \\
 & \multirow[c]{2}{*}{\texttt{vm}} & 6.71e-08 & \bfseries 3.12e-08 & 5.13e-03 & 3.73e-06 & \bfseries 1.90e-06 & 4.57e-03 & 1.65e-05 & \bfseries 1.50e-05 & 5.03e-03 \\
 &  & {\cellcolor{lightgray}} 1.45e-06 & \bfseries {\cellcolor{lightgray}} 1.10e-06 & {\cellcolor{lightgray}} 5.12e-03 & {\cellcolor{lightgray}} 6.07e-06 & \bfseries {\cellcolor{lightgray}} 2.90e-06 & {\cellcolor{lightgray}} 4.57e-03 & {\cellcolor{lightgray}} 1.84e-05 & \bfseries {\cellcolor{lightgray}} 1.60e-05 & {\cellcolor{lightgray}} 5.03e-03 \\
\multirow[c]{4}{*}{Gen} & \multirow[c]{2}{*}{\texttt{pg}} & 1.63e-04 & \bfseries 5.79e-05 & 6.80e-03 & 3.64e-04 & \bfseries 3.17e-04 & 1.41e-02 & 4.88e-04 & \bfseries 4.77e-04 & 1.99e-02 \\
 &  & {\cellcolor{lightgray}} 6.65e-04 & \bfseries {\cellcolor{lightgray}} 2.96e-04 & {\cellcolor{lightgray}} 6.74e-03 & {\cellcolor{lightgray}} 5.28e-04 & \bfseries {\cellcolor{lightgray}} 4.96e-04 & {\cellcolor{lightgray}} 1.41e-02 & \bfseries {\cellcolor{lightgray}} 5.41e-04 & {\cellcolor{lightgray}} 5.62e-04 & {\cellcolor{lightgray}} 1.99e-02 \\
 & \multirow[c]{2}{*}{\texttt{qg}} & 1.88e-05 & \bfseries 1.01e-05 & - & 9.67e-04 & \bfseries 7.88e-04 & - & 1.29e-03 & \bfseries 1.23e-03 & - \\
 &  & {\cellcolor{lightgray}} 1.79e-04 & \bfseries {\cellcolor{lightgray}} 1.22e-04 & {\cellcolor{lightgray}} - & {\cellcolor{lightgray}} 1.39e-03 & \bfseries {\cellcolor{lightgray}} 1.05e-03 & {\cellcolor{lightgray}} - & {\cellcolor{lightgray}} 1.40e-03 & \bfseries {\cellcolor{lightgray}} 1.30e-03 & {\cellcolor{lightgray}} - \\
\multirow[c]{8}{*}{Line } & \multirow[c]{2}{*}{\texttt{pf}} & 5.02e-04 & \bfseries 1.10e-04 & 6.28e-03 & 5.46e-04 & \bfseries 3.67e-04 & 1.70e-02 & \bfseries 1.25e-03 & 1.28e-03 & 1.40e-02 \\
 &  & {\cellcolor{lightgray}} 2.02e-03 & \bfseries {\cellcolor{lightgray}} 1.11e-03 & {\cellcolor{lightgray}} 6.32e-03 & {\cellcolor{lightgray}} 8.67e-04 & \bfseries {\cellcolor{lightgray}} 7.04e-04 & {\cellcolor{lightgray}} 1.69e-02 & \bfseries {\cellcolor{lightgray}} 1.38e-03 & {\cellcolor{lightgray}} 1.49e-03 & {\cellcolor{lightgray}} 1.40e-02 \\
 & \multirow[c]{2}{*}{\texttt{pt}} & 5.03e-04 & \bfseries 1.10e-04 & 6.38e-03 & 5.45e-04 & \bfseries 3.66e-04 & 1.71e-02 & \bfseries 1.25e-03 & 1.28e-03 & 1.43e-02 \\
 &  & {\cellcolor{lightgray}} 2.01e-03 & \bfseries {\cellcolor{lightgray}} 1.11e-03 & {\cellcolor{lightgray}} 6.42e-03 & {\cellcolor{lightgray}} 8.65e-04 & \bfseries {\cellcolor{lightgray}} 7.02e-04 & {\cellcolor{lightgray}} 1.71e-02 & \bfseries {\cellcolor{lightgray}} 1.38e-03 & {\cellcolor{lightgray}} 1.49e-03 & {\cellcolor{lightgray}} 1.43e-02 \\
 & \multirow[c]{2}{*}{\texttt{qf}} & 2.58e-05 & \bfseries 1.21e-05 & - & 4.00e-04 & \bfseries 3.04e-04 & - & 1.28e-03 & \bfseries 1.17e-03 & - \\
 &  & {\cellcolor{lightgray}} 4.43e-03 & \bfseries {\cellcolor{lightgray}} 2.42e-04 & {\cellcolor{lightgray}} - & {\cellcolor{lightgray}} 6.17e-04 & \bfseries {\cellcolor{lightgray}} 3.88e-04 & {\cellcolor{lightgray}} - & {\cellcolor{lightgray}} 1.30e-03 & \bfseries {\cellcolor{lightgray}} 1.19e-03 & {\cellcolor{lightgray}} - \\
 & \multirow[c]{2}{*}{\texttt{qt}} & 2.56e-05 & \bfseries 1.21e-05 & - & 4.01e-04 & \bfseries 3.05e-04 & - & 1.28e-03 & \bfseries 1.17e-03 & - \\
 &  & {\cellcolor{lightgray}} 3.99e-03 & \bfseries {\cellcolor{lightgray}} 2.44e-04 & {\cellcolor{lightgray}} - & {\cellcolor{lightgray}} 6.18e-04 & \bfseries {\cellcolor{lightgray}} 3.90e-04 & {\cellcolor{lightgray}} - & {\cellcolor{lightgray}} 1.30e-03 & \bfseries {\cellcolor{lightgray}} 1.19e-03 & {\cellcolor{lightgray}} - \\
\multirow[c]{8}{*}{Transf.} & \multirow[c]{2}{*}{\texttt{pf}} & 2.30e-04 & \bfseries 6.47e-05 & 7.22e-03 & 1.81e-04 & \bfseries 1.44e-04 & 4.28e-03 & 7.79e-04 & \bfseries 7.53e-04 & 2.25e-02 \\
 &  & {\cellcolor{lightgray}} 1.22e-03 & \bfseries {\cellcolor{lightgray}} 6.17e-04 & {\cellcolor{lightgray}} 7.16e-03 & {\cellcolor{lightgray}} 2.87e-04 & \bfseries {\cellcolor{lightgray}} 2.44e-04 & {\cellcolor{lightgray}} 4.28e-03 & {\cellcolor{lightgray}} 9.57e-04 & \bfseries {\cellcolor{lightgray}} 9.21e-04 & {\cellcolor{lightgray}} 2.26e-02 \\
 & \multirow[c]{2}{*}{\texttt{pt}} & 2.31e-04 & \bfseries 6.49e-05 & 7.30e-03 & 1.81e-04 & \bfseries 1.44e-04 & 4.32e-03 & 7.82e-04 & \bfseries 7.56e-04 & 2.26e-02 \\
 &  & {\cellcolor{lightgray}} 1.22e-03 & \bfseries {\cellcolor{lightgray}} 6.19e-04 & {\cellcolor{lightgray}} 7.24e-03 & {\cellcolor{lightgray}} 2.87e-04 & \bfseries {\cellcolor{lightgray}} 2.45e-04 & {\cellcolor{lightgray}} 4.31e-03 & {\cellcolor{lightgray}} 9.60e-04 & \bfseries {\cellcolor{lightgray}} 9.24e-04 & {\cellcolor{lightgray}} 2.26e-02 \\
 & \multirow[c]{2}{*}{\texttt{qf}} & 1.98e-05 & \bfseries 9.61e-06 & - & 2.76e-04 & \bfseries 2.40e-04 & - & 1.38e-03 & \bfseries 1.30e-03 & - \\
 &  & {\cellcolor{lightgray}} 2.21e-04 & \bfseries {\cellcolor{lightgray}} 1.46e-04 & {\cellcolor{lightgray}} - & {\cellcolor{lightgray}} 3.92e-04 & \bfseries {\cellcolor{lightgray}} 2.84e-04 & {\cellcolor{lightgray}} - & {\cellcolor{lightgray}} 1.46e-03 & \bfseries {\cellcolor{lightgray}} 1.39e-03 & {\cellcolor{lightgray}} - \\
 & \multirow[c]{2}{*}{\texttt{qt}} & 1.94e-05 & \bfseries 9.44e-06 & - & 2.78e-04 & \bfseries 2.42e-04 & - & 1.40e-03 & \bfseries 1.32e-03 & - \\
 &  & {\cellcolor{lightgray}} 2.23e-04 & \bfseries {\cellcolor{lightgray}} 1.46e-04 & {\cellcolor{lightgray}} - & {\cellcolor{lightgray}} 3.96e-04 & \bfseries {\cellcolor{lightgray}} 2.88e-04 & {\cellcolor{lightgray}} - & {\cellcolor{lightgray}} 1.47e-03 & \bfseries {\cellcolor{lightgray}} 1.41e-03 & {\cellcolor{lightgray}} - \\
\hline
\end{tabular}
}
\caption{Pre-power flow MSE of \textsc{DC-IPOPT} and  \textsc{CANOS} across different grid sizes. The non-shaded rows are \textsc{FullTop} results, while the shaded rows are \textsc{TopDrop} results. For each grid and each predicted value, we highlight the best performing model architecture in bold.}
\label{tab:pre_pf_mse}
\end{table}

\begin{table}[!h]
\centering
\resizebox{\textwidth}{!}{\begin{tabular}{llrrr|rrr|rrr}
\hline
 &  & \multicolumn{3}{c|}{500} & \multicolumn{3}{c|}{2000} & \multicolumn{3}{c}{10000} \\
 &  & \textsc{Wide-CANOS} & \textsc{Deep-CANOS}$_{48}$ & \textsc{DC-IPOPT} & \textsc{Deep-CANOS}$_{48}$ & \textsc{Deep-CANOS}$_{60}$ & \textsc{DC-IPOPT} & \textsc{Deep-CANOS}$_{48}$ & \textsc{Deep-CANOS}$_{60}$ & \textsc{DC-IPOPT} \\
\hline
\multirow[c]{4}{*}{Bus } & \multirow[c]{2}{*}{\texttt{va}} & 7.69e-05 & \bfseries 1.24e-05 & 3.00e-02 & 9.69e-04 & \bfseries 8.23e-04 & 2.14e-01 & 6.01e-04 & \bfseries 5.81e-04 & 1.55e-01 \\
 &  & {\cellcolor{lightgray}} 4.02e-04 & \bfseries {\cellcolor{lightgray}} 1.74e-04 & {\cellcolor{lightgray}} 2.99e-02 & {\cellcolor{lightgray}} 1.59e-03 & \bfseries {\cellcolor{lightgray}} 1.39e-03 & {\cellcolor{lightgray}} 2.14e-01 & \bfseries {\cellcolor{lightgray}} 6.00e-04 & {\cellcolor{lightgray}} 6.09e-04 & {\cellcolor{lightgray}} 1.55e-01 \\
 & \multirow[c]{2}{*}{\texttt{vm}} & 7.74e-08 & \bfseries 3.36e-08 & 7.00e-03 & 6.27e-06 & \bfseries 2.66e-06 & 5.97e-03 & 1.72e-05 & \bfseries 1.62e-05 & 4.60e-03 \\
 &  & {\cellcolor{lightgray}} 1.30e-06 & \bfseries {\cellcolor{lightgray}} 1.10e-06 & {\cellcolor{lightgray}} 7.00e-03 & {\cellcolor{lightgray}} 1.04e-05 & \bfseries {\cellcolor{lightgray}} 4.46e-06 & {\cellcolor{lightgray}} 5.98e-03 & {\cellcolor{lightgray}} 1.80e-05 & \bfseries {\cellcolor{lightgray}} 1.56e-05 & {\cellcolor{lightgray}} 4.60e-03 \\
\multirow[c]{4}{*}{Gen} & \multirow[c]{2}{*}{\texttt{pg}} & 7.70e-04 & \bfseries 2.46e-04 & 1.02e-01 & 6.01e-03 & \bfseries 5.24e-03 & 5.03e-01 & 1.61e-03 & \bfseries 1.57e-03 & 2.40e-01 \\
 &  & {\cellcolor{lightgray}} 3.89e-03 & \bfseries {\cellcolor{lightgray}} 1.47e-03 & {\cellcolor{lightgray}} 1.02e-01 & {\cellcolor{lightgray}} 1.01e-02 & \bfseries {\cellcolor{lightgray}} 9.09e-03 & {\cellcolor{lightgray}} 5.03e-01 & \bfseries {\cellcolor{lightgray}} 1.66e-03 & {\cellcolor{lightgray}} 1.69e-03 & {\cellcolor{lightgray}} 2.40e-01 \\
 & \multirow[c]{2}{*}{\texttt{qg}} & 3.54e-05 & \bfseries 1.44e-05 & 1.82e-01 & 1.22e-03 & \bfseries 1.02e-03 & 3.10e-01 & 1.39e-03 & \bfseries 1.35e-03 & 1.11e-01 \\
 &  & {\cellcolor{lightgray}} 2.58e-04 & \bfseries {\cellcolor{lightgray}} 1.65e-04 & {\cellcolor{lightgray}} 1.82e-01 & {\cellcolor{lightgray}} 1.91e-03 & \bfseries {\cellcolor{lightgray}} 1.39e-03 & {\cellcolor{lightgray}} 3.10e-01 & {\cellcolor{lightgray}} 1.39e-03 & \bfseries {\cellcolor{lightgray}} 1.32e-03 & {\cellcolor{lightgray}} 1.11e-01 \\
\multirow[c]{8}{*}{Line } & \multirow[c]{2}{*}{\texttt{pf}} & 3.37e-04 & \bfseries 7.16e-05 & 4.27e-02 & 1.27e-03 & \bfseries 1.10e-03 & 7.68e-02 & 3.81e-04 & \bfseries 3.72e-04 & 6.86e-02 \\
 &  & {\cellcolor{lightgray}} 1.51e-03 & \bfseries {\cellcolor{lightgray}} 6.53e-04 & {\cellcolor{lightgray}} 4.30e-02 & {\cellcolor{lightgray}} 2.04e-03 & \bfseries {\cellcolor{lightgray}} 1.81e-03 & {\cellcolor{lightgray}} 7.66e-02 & \bfseries {\cellcolor{lightgray}} 3.90e-04 & {\cellcolor{lightgray}} 3.99e-04 & {\cellcolor{lightgray}} 6.87e-02 \\
 & \multirow[c]{2}{*}{\texttt{pt}} & 3.37e-04 & \bfseries 7.13e-05 & 3.17e-02 & 1.26e-03 & \bfseries 1.09e-03 & 1.12e-01 & 3.82e-04 & \bfseries 3.73e-04 & 5.51e-02 \\
 &  & {\cellcolor{lightgray}} 1.51e-03 & \bfseries {\cellcolor{lightgray}} 6.53e-04 & {\cellcolor{lightgray}} 3.17e-02 & {\cellcolor{lightgray}} 2.02e-03 & \bfseries {\cellcolor{lightgray}} 1.80e-03 & {\cellcolor{lightgray}} 1.12e-01 & \bfseries {\cellcolor{lightgray}} 3.91e-04 & {\cellcolor{lightgray}} 3.99e-04 & {\cellcolor{lightgray}} 5.51e-02 \\
 & \multirow[c]{2}{*}{\texttt{qf}} & 1.62e-05 & \bfseries 6.66e-06 & 3.20e-02 & 3.71e-04 & \bfseries 2.88e-04 & 1.14e-01 & 1.05e-03 & \bfseries 1.02e-03 & 5.52e-02 \\
 &  & {\cellcolor{lightgray}} 1.29e-04 & \bfseries {\cellcolor{lightgray}} 9.70e-05 & {\cellcolor{lightgray}} 3.20e-02 & {\cellcolor{lightgray}} 5.38e-04 & \bfseries {\cellcolor{lightgray}} 3.68e-04 & {\cellcolor{lightgray}} 1.14e-01 & {\cellcolor{lightgray}} 1.09e-03 & \bfseries {\cellcolor{lightgray}} 9.81e-04 & {\cellcolor{lightgray}} 5.52e-02 \\
 & \multirow[c]{2}{*}{\texttt{qt}} & 1.74e-05 & \bfseries 6.99e-06 & 3.82e-02 & 3.81e-04 & \bfseries 2.97e-04 & 7.52e-02 & 1.05e-03 & \bfseries 1.02e-03 & 6.84e-02 \\
 &  & {\cellcolor{lightgray}} 1.35e-04 & \bfseries {\cellcolor{lightgray}} 9.97e-05 & {\cellcolor{lightgray}} 3.86e-02 & {\cellcolor{lightgray}} 5.52e-04 & \bfseries {\cellcolor{lightgray}} 3.81e-04 & {\cellcolor{lightgray}} 7.50e-02 & {\cellcolor{lightgray}} 1.09e-03 & \bfseries {\cellcolor{lightgray}} 9.80e-04 & {\cellcolor{lightgray}} 6.85e-02 \\
\multirow[c]{8}{*}{Transf.} & \multirow[c]{2}{*}{\texttt{pf}} & 7.18e-04 & \bfseries 2.24e-04 & 1.72e-01 & 1.65e-03 & \bfseries 1.44e-03 & 8.29e-02 & 1.43e-03 & \bfseries 1.40e-03 & 1.05e-01 \\
 &  & {\cellcolor{lightgray}} 3.63e-03 & \bfseries {\cellcolor{lightgray}} 1.37e-03 & {\cellcolor{lightgray}} 1.72e-01 & {\cellcolor{lightgray}} 2.76e-03 & \bfseries {\cellcolor{lightgray}} 2.49e-03 & {\cellcolor{lightgray}} 8.27e-02 & \bfseries {\cellcolor{lightgray}} 1.48e-03 & {\cellcolor{lightgray}} 1.51e-03 & {\cellcolor{lightgray}} 1.05e-01 \\
 & \multirow[c]{2}{*}{\texttt{pt}} & 7.21e-04 & \bfseries 2.25e-04 & 9.36e-02 & 1.66e-03 & \bfseries 1.45e-03 & 1.38e-01 & 1.44e-03 & \bfseries 1.41e-03 & 2.14e-01 \\
 &  & {\cellcolor{lightgray}} 3.64e-03 & \bfseries {\cellcolor{lightgray}} 1.38e-03 & {\cellcolor{lightgray}} 9.36e-02 & {\cellcolor{lightgray}} 2.78e-03 & \bfseries {\cellcolor{lightgray}} 2.51e-03 & {\cellcolor{lightgray}} 1.38e-01 & \bfseries {\cellcolor{lightgray}} 1.49e-03 & {\cellcolor{lightgray}} 1.51e-03 & {\cellcolor{lightgray}} 2.14e-01 \\
 & \multirow[c]{2}{*}{\texttt{qf}} & 2.22e-05 & \bfseries 9.82e-06 & 9.29e-02 & 2.47e-04 & \bfseries 2.10e-04 & 1.37e-01 & 1.30e-03 & \bfseries 1.26e-03 & 2.12e-01 \\
 &  & {\cellcolor{lightgray}} 1.78e-04 & \bfseries {\cellcolor{lightgray}} 1.24e-04 & {\cellcolor{lightgray}} 9.30e-02 & {\cellcolor{lightgray}} 3.72e-04 & \bfseries {\cellcolor{lightgray}} 2.57e-04 & {\cellcolor{lightgray}} 1.37e-01 & {\cellcolor{lightgray}} 1.31e-03 & \bfseries {\cellcolor{lightgray}} 1.23e-03 & {\cellcolor{lightgray}} 2.12e-01 \\
 & \multirow[c]{2}{*}{\texttt{qt}} & 2.52e-05 & \bfseries 1.03e-05 & 1.52e-01 & 2.74e-04 & \bfseries 2.33e-04 & 7.90e-02 & 1.34e-03 & \bfseries 1.29e-03 & 9.31e-02 \\
 &  & {\cellcolor{lightgray}} 1.99e-04 & \bfseries {\cellcolor{lightgray}} 1.32e-04 & {\cellcolor{lightgray}} 1.52e-01 & {\cellcolor{lightgray}} 4.37e-04 & \bfseries {\cellcolor{lightgray}} 3.16e-04 & {\cellcolor{lightgray}} 7.87e-02 & {\cellcolor{lightgray}} 1.35e-03 & \bfseries {\cellcolor{lightgray}} 1.26e-03 & {\cellcolor{lightgray}} 9.32e-02 \\
\hline

\end{tabular}
}
\caption{Post-power flow MSE of \textsc{DC-IPOPT} and  \textsc{CANOS} across different grid sizes. The non-shaded rows are \textsc{FullTop} results, while the shaded rows are \textsc{TopDrop} results. For each grid and each predicted value, we highlight the best performing model architecture in bold.}
\label{tab:post_pf_mse}
\end{table}

\clearpage
\subsection{Feasibility at threshold - Pre-power flow}
In this section we report the percentage of entities which satisfy constraints at different thresholds values ($0.01$, $0.0001$, $10^{-6}$, $10^{-8}$), before power flow. Note that the threshold refers to the absolute violation (constraint violation degree, see Section \ref{sec:training}).

\begin{table}[!h]
\centering
\resizebox{.95\textwidth}{!}{\begin{tabular}{lrr|rr|rr}
\hline
 & \multicolumn{2}{c|}{500} & \multicolumn{2}{c|}{2000} & \multicolumn{2}{c}{10000} \\
 & \textsc{Wide-CANOS} & \textsc{Deep-CANOS}$_{48}$ & \textsc{Deep-CANOS}$_{48}$ & \textsc{Deep-CANOS}$_{60}$ & \textsc{Deep-CANOS}$_{48}$ & \textsc{Deep-CANOS}$_{60}$ \\
\hline
\multirow[c]{2}{*}{Branch thermal limit from} & 100.00 \% & 100.00 \% & 100.00 \% & 100.00 \% & 100.00 \% & 100.00 \% \\
 & {\cellcolor{lightgray}} 99.99 \% & {\cellcolor{lightgray}} 99.99 \% & {\cellcolor{lightgray}} 100.00 \% & {\cellcolor{lightgray}} 100.00 \% & {\cellcolor{lightgray}} 100.00 \% & {\cellcolor{lightgray}} 100.00 \% \\
\multirow[c]{2}{*}{Branch thermal limit to} & 100.00 \% & 100.00 \% & 100.00 \% & 100.00 \% & 100.00 \% & 100.00 \% \\
 & {\cellcolor{lightgray}} 100.00 \% & {\cellcolor{lightgray}} 100.00 \% & {\cellcolor{lightgray}} 100.00 \% & {\cellcolor{lightgray}} 100.00 \% & {\cellcolor{lightgray}} 100.00 \% & {\cellcolor{lightgray}} 100.00 \% \\
\multirow[c]{2}{*}{Branch voltage angle difference} & 100.00 \% & 100.00 \% & 100.00 \% & 100.00 \% & 100.00 \% & 100.00 \% \\
 & {\cellcolor{lightgray}} 100.00 \% & {\cellcolor{lightgray}} 100.00 \% & {\cellcolor{lightgray}} 100.00 \% & {\cellcolor{lightgray}} 100.00 \% & {\cellcolor{lightgray}} 100.00 \% & {\cellcolor{lightgray}} 100.00 \% \\
\multirow[c]{2}{*}{Bus voltage bounds} & 100.00 \% & 100.00 \% & 100.00 \% & 100.00 \% & 100.00 \% & 100.00 \% \\
 & {\cellcolor{lightgray}} 100.00 \% & {\cellcolor{lightgray}} 100.00 \% & {\cellcolor{lightgray}} 100.00 \% & {\cellcolor{lightgray}} 100.00 \% & {\cellcolor{lightgray}} 100.00 \% & {\cellcolor{lightgray}} 100.00 \% \\
\multirow[c]{2}{*}{Reactive power balance bus} & 95.75 \% & 96.83 \% & 84.75 \% & 89.06 \% & 72.11 \% & 70.66 \% \\
 & {\cellcolor{lightgray}} 90.66 \% & {\cellcolor{lightgray}} 91.35 \% & {\cellcolor{lightgray}} 87.10 \% & {\cellcolor{lightgray}} 86.97 \% & {\cellcolor{lightgray}} 72.62 \% & {\cellcolor{lightgray}} 71.40 \% \\
\multirow[c]{2}{*}{Real power balance bus} & 81.51 \% & 85.98 \% & 69.89 \% & 73.10 \% & 47.80 \% & 47.00 \% \\
 & {\cellcolor{lightgray}} 71.73 \% & {\cellcolor{lightgray}} 73.89 \% & {\cellcolor{lightgray}} 70.39 \% & {\cellcolor{lightgray}} 71.85 \% & {\cellcolor{lightgray}} 49.66 \% & {\cellcolor{lightgray}} 46.52 \% \\
\hline

\end{tabular}}
\caption{Pre-power flow percentage of entities \textbf{below} $\mathbf{0.01}$ constraint violation. The non-shaded rows are \textsc{FullTop} results, while the shaded rows are \textsc{TopDrop} results. For each grid and each predicted value, we highlight the best performing model architecture in bold.}
\label{tab:pre_0.01}
\end{table}

\begin{table}[!h]
\centering
\resizebox{.95\textwidth}{!}{\begin{tabular}{lrr|rr|rr}
\hline
 & \multicolumn{2}{c|}{500} & \multicolumn{2}{c|}{2000} & \multicolumn{2}{c}{10000} \\
 & \textsc{Wide-CANOS} & \textsc{Deep-CANOS}$_{48}$ & \textsc{Deep-CANOS}$_{48}$ & \textsc{Deep-CANOS}$_{60}$ & \textsc{Deep-CANOS}$_{48}$ & \textsc{Deep-CANOS}$_{60}$ \\
\hline
\multirow[c]{2}{*}{Branch thermal limit from} & 100.00 \% & 100.00 \% & 100.00 \% & 100.00 \% & 100.00 \% & 100.00 \% \\
 & {\cellcolor{lightgray}} 99.99 \% & {\cellcolor{lightgray}} 99.99 \% & {\cellcolor{lightgray}} 100.00 \% & {\cellcolor{lightgray}} 100.00 \% & {\cellcolor{lightgray}} 100.00 \% & {\cellcolor{lightgray}} 100.00 \% \\
\multirow[c]{2}{*}{Branch thermal limit to} & 100.00 \% & 100.00 \% & 99.99 \% & 99.99 \% & 100.00 \% & 100.00 \% \\
 & {\cellcolor{lightgray}} 100.00 \% & {\cellcolor{lightgray}} 100.00 \% & {\cellcolor{lightgray}} 99.99 \% & {\cellcolor{lightgray}} 99.99 \% & {\cellcolor{lightgray}} 100.00 \% & {\cellcolor{lightgray}} 100.00 \% \\
\multirow[c]{2}{*}{Branch voltage angle difference} & 100.00 \% & 100.00 \% & 100.00 \% & 100.00 \% & 100.00 \% & 100.00 \% \\
 & {\cellcolor{lightgray}} 100.00 \% & {\cellcolor{lightgray}} 100.00 \% & {\cellcolor{lightgray}} 100.00 \% & {\cellcolor{lightgray}} 100.00 \% & {\cellcolor{lightgray}} 100.00 \% & {\cellcolor{lightgray}} 100.00 \% \\
\multirow[c]{2}{*}{Bus voltage bounds} & 100.00 \% & 100.00 \% & 100.00 \% & 100.00 \% & 100.00 \% & 100.00 \% \\
 & {\cellcolor{lightgray}} 100.00 \% & {\cellcolor{lightgray}} 100.00 \% & {\cellcolor{lightgray}} 100.00 \% & {\cellcolor{lightgray}} 100.00 \% & {\cellcolor{lightgray}} 100.00 \% & {\cellcolor{lightgray}} 100.00 \% \\
\multirow[c]{2}{*}{Reactive power balance bus} & 7.68 \% & 9.29 \% & 2.85 \% & 3.39 \% & 2.98 \% & 2.78 \% \\
 & {\cellcolor{lightgray}} 3.85 \% & {\cellcolor{lightgray}} 4.14 \% & {\cellcolor{lightgray}} 3.29 \% & {\cellcolor{lightgray}} 3.01 \% & {\cellcolor{lightgray}} 2.95 \% & {\cellcolor{lightgray}} 2.65 \% \\
\multirow[c]{2}{*}{Real power balance bus} & 2.92 \% & 4.35 \% & 1.61 \% & 2.31 \% & 1.41 \% & 1.40 \% \\
 & {\cellcolor{lightgray}} 1.89 \% & {\cellcolor{lightgray}} 1.91 \% & {\cellcolor{lightgray}} 2.00 \% & {\cellcolor{lightgray}} 2.10 \% & {\cellcolor{lightgray}} 1.57 \% & {\cellcolor{lightgray}} 1.30 \% \\
\hline
\end{tabular}}
\caption{Pre-power flow percentage of entities below \textbf{below} $\mathbf{0.0001}$ constraint violation. The non-shaded rows are \textsc{FullTop} results, while the shaded rows are \textsc{TopDrop} results.}
\label{tab:pre_0.0001}
\end{table}

\begin{table}[!h]
\centering
\resizebox{.95\textwidth}{!}{\begin{tabular}{lrr|rr|rr}
\hline
 & \multicolumn{2}{c|}{500} & \multicolumn{2}{c|}{2000} & \multicolumn{2}{c}{10000} \\
 & \textsc{Wide-CANOS} & \textsc{Deep-CANOS}$_{48}$ & \textsc{Deep-CANOS}$_{48}$ & \textsc{Deep-CANOS}$_{60}$ & \textsc{Deep-CANOS}$_{48}$ & \textsc{Deep-CANOS}$_{60}$ \\
\hline
\multirow[c]{2}{*}{Branch thermal limit from} & 100.00 \% & 100.00 \% & 100.00 \% & 100.00 \% & 100.00 \% & 100.00 \% \\
 & {\cellcolor{lightgray}} 99.99 \% & {\cellcolor{lightgray}} 99.99 \% & {\cellcolor{lightgray}} 100.00 \% & {\cellcolor{lightgray}} 100.00 \% & {\cellcolor{lightgray}} 100.00 \% & {\cellcolor{lightgray}} 100.00 \% \\
\multirow[c]{2}{*}{Branch thermal limit to} & 100.00 \% & 100.00 \% & 99.99 \% & 99.99 \% & 100.00 \% & 100.00 \% \\
 & {\cellcolor{lightgray}} 100.00 \% & {\cellcolor{lightgray}} 100.00 \% & {\cellcolor{lightgray}} 99.99 \% & {\cellcolor{lightgray}} 99.99 \% & {\cellcolor{lightgray}} 100.00 \% & {\cellcolor{lightgray}} 100.00 \% \\
\multirow[c]{2}{*}{Branch voltage angle difference} & 100.00 \% & 100.00 \% & 100.00 \% & 100.00 \% & 100.00 \% & 100.00 \% \\
 & {\cellcolor{lightgray}} 100.00 \% & {\cellcolor{lightgray}} 100.00 \% & {\cellcolor{lightgray}} 100.00 \% & {\cellcolor{lightgray}} 100.00 \% & {\cellcolor{lightgray}} 100.00 \% & {\cellcolor{lightgray}} 100.00 \% \\
\multirow[c]{2}{*}{Bus voltage bounds} & 100.00 \% & 100.00 \% & 100.00 \% & 100.00 \% & 100.00 \% & 100.00 \% \\
 & {\cellcolor{lightgray}} 100.00 \% & {\cellcolor{lightgray}} 100.00 \% & {\cellcolor{lightgray}} 100.00 \% & {\cellcolor{lightgray}} 100.00 \% & {\cellcolor{lightgray}} 100.00 \% & {\cellcolor{lightgray}} 100.00 \% \\
\multirow[c]{2}{*}{Reactive power balance bus} & 0.09 \% & 0.11 \% & 0.03 \% & 0.03 \% & 0.03 \% & 0.04 \% \\
 & {\cellcolor{lightgray}} 0.04 \% & {\cellcolor{lightgray}} 0.05 \% & {\cellcolor{lightgray}} 0.03 \% & {\cellcolor{lightgray}} 0.03 \% & {\cellcolor{lightgray}} 0.04 \% & {\cellcolor{lightgray}} 0.03 \% \\
\multirow[c]{2}{*}{Real power balance bus} & 0.03 \% & 0.05 \% & 0.02 \% & 0.02 \% & 0.02 \% & 0.02 \% \\
 & {\cellcolor{lightgray}} 0.02 \% & {\cellcolor{lightgray}} 0.02 \% & {\cellcolor{lightgray}} 0.02 \% & {\cellcolor{lightgray}} 0.02 \% & {\cellcolor{lightgray}} 0.02 \% & {\cellcolor{lightgray}} 0.01 \% \\
\hline
\end{tabular}
}
\caption{Pre-power flow percentage of entities \textbf{below} $\mathbf{10^{-6}}$ constraint violation. The non-shaded rows are \textsc{FullTop} results, while the shaded rows are \textsc{TopDrop} results.}
\label{tab:pre_1e-06}
\end{table}

\begin{table}[!h]
\centering
\resizebox{.95\textwidth}{!}{\begin{tabular}{lrr|rr|rr}
\hline
 & \multicolumn{2}{c|}{500} & \multicolumn{2}{c|}{2000} & \multicolumn{2}{c}{10000} \\
 & \textsc{Wide-CANOS} & \textsc{Deep-CANOS}$_{48}$ & \textsc{Deep-CANOS}$_{48}$ & \textsc{Deep-CANOS}$_{60}$ & \textsc{Deep-CANOS}$_{48}$ & \textsc{Deep-CANOS}$_{60}$ \\
\hline
\multirow[c]{2}{*}{Branch thermal limit from} & 100.00 \% & 100.00 \% & 100.00 \% & 100.00 \% & 100.00 \% & 100.00 \% \\
 & {\cellcolor{lightgray}} 99.99 \% & {\cellcolor{lightgray}} 99.99 \% & {\cellcolor{lightgray}} 100.00 \% & {\cellcolor{lightgray}} 100.00 \% & {\cellcolor{lightgray}} 100.00 \% & {\cellcolor{lightgray}} 100.00 \% \\
\multirow[c]{2}{*}{Branch thermal limit to} & 100.00 \% & 100.00 \% & 99.99 \% & 99.99 \% & 100.00 \% & 100.00 \% \\
 & {\cellcolor{lightgray}} 100.00 \% & {\cellcolor{lightgray}} 100.00 \% & {\cellcolor{lightgray}} 99.99 \% & {\cellcolor{lightgray}} 99.99 \% & {\cellcolor{lightgray}} 100.00 \% & {\cellcolor{lightgray}} 100.00 \% \\
\multirow[c]{2}{*}{Branch voltage angle difference} & 100.00 \% & 100.00 \% & 100.00 \% & 100.00 \% & 100.00 \% & 100.00 \% \\
 & {\cellcolor{lightgray}} 100.00 \% & {\cellcolor{lightgray}} 100.00 \% & {\cellcolor{lightgray}} 100.00 \% & {\cellcolor{lightgray}} 100.00 \% & {\cellcolor{lightgray}} 100.00 \% & {\cellcolor{lightgray}} 100.00 \% \\
\multirow[c]{2}{*}{Bus voltage bounds} & 100.00 \% & 100.00 \% & 100.00 \% & 100.00 \% & 100.00 \% & 100.00 \% \\
 & {\cellcolor{lightgray}} 100.00 \% & {\cellcolor{lightgray}} 100.00 \% & {\cellcolor{lightgray}} 100.00 \% & {\cellcolor{lightgray}} 100.00 \% & {\cellcolor{lightgray}} 100.00 \% & {\cellcolor{lightgray}} 100.00 \% \\
\multirow[c]{2}{*}{Reactive power balance bus} & 0.00 \% & 0.00 \% & 0.00 \% & 0.00 \% & 0.00 \% & 0.00 \% \\
 & {\cellcolor{lightgray}} 0.00 \% & {\cellcolor{lightgray}} 0.00 \% & {\cellcolor{lightgray}} 0.00 \% & {\cellcolor{lightgray}} 0.00 \% & {\cellcolor{lightgray}} 0.00 \% & {\cellcolor{lightgray}} 0.00 \% \\
\multirow[c]{2}{*}{Real power balance bus} & 0.00 \% & 0.00 \% & 0.00 \% & 0.00 \% & 0.00 \% & 0.00 \% \\
 & {\cellcolor{lightgray}} 0.00 \% & {\cellcolor{lightgray}} 0.00 \% & {\cellcolor{lightgray}} 0.00 \% & {\cellcolor{lightgray}} 0.00 \% & {\cellcolor{lightgray}} 0.00 \% & {\cellcolor{lightgray}} 0.00 \% \\
\hline
\end{tabular}
}
\caption{Pre-power flow percentage of entities \textbf{below} $\mathbf{10^{-8}}$ constraint violation. The non-shaded rows are \textsc{FullTop} results, while the shaded rows are \textsc{TopDrop} results.}
\label{tab:pre_1e-08}
\end{table}

\clearpage
\subsection{Feasibility at threshold - Post-power flow}
In this section we report the percentage of entities which satisfy constraints at different thresholds values ($0.01$, $0.0001$, $10^{-6}$, $10^{-8}$), after power flow. Note that the threshold refers to the absolute violation (constraint violation degree, see Section \ref{sec:training}).

\begin{table}[!h]
\centering
\resizebox{\textwidth}{!}{\begin{tabular}{lrrr|rrr|rrr}
\hline
 & \multicolumn{3}{c|}{500} & \multicolumn{3}{c|}{2000} & \multicolumn{3}{c}{10000} \\
 & \textsc{Wide-CANOS} & \textsc{Deep-CANOS}$_{48}$ & \textsc{DC-IPOPT} & \textsc{Deep-CANOS}$_{48}$ & \textsc{Deep-CANOS}$_{60}$ & \textsc{DC-IPOPT} & \textsc{Deep-CANOS}$_{48}$ & \textsc{Deep-CANOS}$_{60}$ & \textsc{DC-IPOPT} \\
\hline
\multirow[c]{2}{*}{Branch thermal limit from} & 99.982 \% & 99.987 \% & 99.847 \% & 99.993 \% & 99.992 \% & 99.809 \% & 99.994 \% & 99.993 \% & 99.985 \% \\
 & {\cellcolor{lightgray}} 99.947 \% & {\cellcolor{lightgray}} 99.982 \% & {\cellcolor{lightgray}} 99.844 \% & {\cellcolor{lightgray}} 99.991 \% & {\cellcolor{lightgray}} 99.991 \% & {\cellcolor{lightgray}} 99.809 \% & {\cellcolor{lightgray}} 99.993 \% & {\cellcolor{lightgray}} 99.994 \% & {\cellcolor{lightgray}} 99.985 \% \\
\multirow[c]{2}{*}{Branch thermal limit to} & 99.999 \% & 99.999 \% & 99.854 \% & 99.990 \% & 99.990 \% & 99.830 \% & 99.997 \% & 99.997 \% & 99.985 \% \\
 & {\cellcolor{lightgray}} 99.987 \% & {\cellcolor{lightgray}} 99.996 \% & {\cellcolor{lightgray}} 99.851 \% & {\cellcolor{lightgray}} 99.989 \% & {\cellcolor{lightgray}} 99.989 \% & {\cellcolor{lightgray}} 99.829 \% & {\cellcolor{lightgray}} 99.997 \% & {\cellcolor{lightgray}} 99.997 \% & {\cellcolor{lightgray}} 99.985 \% \\
\multirow[c]{2}{*}{Branch voltage angle difference} & 100.000 \% & 100.000 \% & 100.000 \% & 100.000 \% & 100.000 \% & 100.000 \% & 100.000 \% & 100.000 \% & 100.000 \% \\
 & {\cellcolor{lightgray}} 100.000 \% & {\cellcolor{lightgray}} 100.000 \% & {\cellcolor{lightgray}} 100.000 \% & {\cellcolor{lightgray}} 100.000 \% & {\cellcolor{lightgray}} 100.000 \% & {\cellcolor{lightgray}} 100.000 \% & {\cellcolor{lightgray}} 100.000 \% & {\cellcolor{lightgray}} 100.000 \% & {\cellcolor{lightgray}} 100.000 \% \\
\multirow[c]{2}{*}{Bus voltage bounds pq} & 100.000 \% & 100.000 \% & 100.000 \% & 99.990 \% & 99.999 \% & 100.000 \% & 99.997 \% & 99.997 \% & 99.767 \% \\
 & {\cellcolor{lightgray}} 100.000 \% & {\cellcolor{lightgray}} 100.000 \% & {\cellcolor{lightgray}} 99.995 \% & {\cellcolor{lightgray}} 99.988 \% & {\cellcolor{lightgray}} 99.996 \% & {\cellcolor{lightgray}} 99.999 \% & {\cellcolor{lightgray}} 99.996 \% & {\cellcolor{lightgray}} 99.996 \% & {\cellcolor{lightgray}} 99.768 \% \\
\multirow[c]{2}{*}{Bus voltage bounds pv} & 100.000 \% & 100.000 \% & 100.000 \% & 99.998 \% & 100.000 \% & 100.000 \% & 100.000 \% & 100.000 \% & 100.000 \% \\
 & {\cellcolor{lightgray}} 100.000 \% & {\cellcolor{lightgray}} 100.000 \% & {\cellcolor{lightgray}} 99.999 \% & {\cellcolor{lightgray}} 99.993 \% & {\cellcolor{lightgray}} 99.998 \% & {\cellcolor{lightgray}} 100.000 \% & {\cellcolor{lightgray}} 99.999 \% & {\cellcolor{lightgray}} 99.999 \% & {\cellcolor{lightgray}} 99.999 \% \\
\multirow[c]{2}{*}{Generator reactive power bounds slack} & 100.000 \% & 100.000 \% & 100.000 \% & 100.000 \% & 100.000 \% & 100.000 \% & 50.311 \% & 47.108 \% & 100.000 \% \\
 & {\cellcolor{lightgray}} 100.000 \% & {\cellcolor{lightgray}} 100.000 \% & {\cellcolor{lightgray}} 99.871 \% & {\cellcolor{lightgray}} 99.911 \% & {\cellcolor{lightgray}} 99.915 \% & {\cellcolor{lightgray}} 99.887 \% & {\cellcolor{lightgray}} 47.174 \% & {\cellcolor{lightgray}} 46.456 \% & {\cellcolor{lightgray}} 99.973 \% \\
\multirow[c]{2}{*}{Generator real power bounds slack} & 53.606 \% & 38.119 \% & 0.000 \% & 47.971 \% & 44.640 \% & 0.000 \% & 63.630 \% & 65.559 \% & 0.000 \% \\
 & {\cellcolor{lightgray}} 64.765 \% & {\cellcolor{lightgray}} 53.814 \% & {\cellcolor{lightgray}} 0.000 \% & {\cellcolor{lightgray}} 49.450 \% & {\cellcolor{lightgray}} 47.838 \% & {\cellcolor{lightgray}} 0.000 \% & {\cellcolor{lightgray}} 65.649 \% & {\cellcolor{lightgray}} 65.367 \% & {\cellcolor{lightgray}} 0.000 \% \\
\hline
\end{tabular}
}
\caption{Post-power flow percentage of entities \textbf{below} $\mathbf{0.01}$ constraint violation. The non-shaded rows are \textsc{FullTop} results, while the shaded rows are \textsc{TopDrop} results.}
\label{tab:post_0.01}
\end{table}

\begin{table}[!h]
\centering
\resizebox{\textwidth}{!}{\begin{tabular}{lrrr|rrr|rrr}
\hline
 & \multicolumn{3}{c|}{500} & \multicolumn{3}{c|}{2000} & \multicolumn{3}{c}{10000} \\
 & \textsc{Wide-CANOS} & \textsc{Deep-CANOS}$_{48}$ & \textsc{DC-IPOPT} & \textsc{Deep-CANOS}$_{48}$ & \textsc{Deep-CANOS}$_{60}$ & \textsc{DC-IPOPT} & \textsc{Deep-CANOS}$_{48}$ & \textsc{Deep-CANOS}$_{60}$ & \textsc{DC-IPOPT} \\
\hline
\multirow[c]{2}{*}{Branch thermal limit from} & 99.974 \% & 99.981 \% & 99.847 \% & 99.989 \% & 99.988 \% & 99.807 \% & 99.989 \% & 99.988 \% & 99.985 \% \\
 & {\cellcolor{lightgray}} 99.941 \% & {\cellcolor{lightgray}} 99.977 \% & {\cellcolor{lightgray}} 99.843 \% & {\cellcolor{lightgray}} 99.988 \% & {\cellcolor{lightgray}} 99.988 \% & {\cellcolor{lightgray}} 99.808 \% & {\cellcolor{lightgray}} 99.989 \% & {\cellcolor{lightgray}} 99.989 \% & {\cellcolor{lightgray}} 99.985 \% \\
\multirow[c]{2}{*}{Branch thermal limit to} & 99.999 \% & 99.999 \% & 99.853 \% & 99.986 \% & 99.985 \% & 99.820 \% & 99.992 \% & 99.992 \% & 99.985 \% \\
 & {\cellcolor{lightgray}} 99.984 \% & {\cellcolor{lightgray}} 99.995 \% & {\cellcolor{lightgray}} 99.851 \% & {\cellcolor{lightgray}} 99.985 \% & {\cellcolor{lightgray}} 99.985 \% & {\cellcolor{lightgray}} 99.820 \% & {\cellcolor{lightgray}} 99.992 \% & {\cellcolor{lightgray}} 99.991 \% & {\cellcolor{lightgray}} 99.985 \% \\
\multirow[c]{2}{*}{Branch voltage angle difference} & 100.000 \% & 100.000 \% & 100.000 \% & 100.000 \% & 100.000 \% & 100.000 \% & 100.000 \% & 100.000 \% & 100.000 \% \\
 & {\cellcolor{lightgray}} 100.000 \% & {\cellcolor{lightgray}} 100.000 \% & {\cellcolor{lightgray}} 100.000 \% & {\cellcolor{lightgray}} 100.000 \% & {\cellcolor{lightgray}} 100.000 \% & {\cellcolor{lightgray}} 100.000 \% & {\cellcolor{lightgray}} 100.000 \% & {\cellcolor{lightgray}} 100.000 \% & {\cellcolor{lightgray}} 100.000 \% \\
\multirow[c]{2}{*}{Bus voltage bounds pq} & 99.566 \% & 99.848 \% & 100.000 \% & 99.725 \% & 99.795 \% & 100.000 \% & 99.782 \% & 99.809 \% & 99.599 \% \\
 & {\cellcolor{lightgray}} 99.186 \% & {\cellcolor{lightgray}} 99.401 \% & {\cellcolor{lightgray}} 99.992 \% & {\cellcolor{lightgray}} 99.689 \% & {\cellcolor{lightgray}} 99.778 \% & {\cellcolor{lightgray}} 99.999 \% & {\cellcolor{lightgray}} 99.778 \% & {\cellcolor{lightgray}} 99.787 \% & {\cellcolor{lightgray}} 99.601 \% \\
\multirow[c]{2}{*}{Bus voltage bounds pv} & 100.000 \% & 100.000 \% & 100.000 \% & 99.924 \% & 99.986 \% & 100.000 \% & 99.877 \% & 99.883 \% & 100.000 \% \\
 & {\cellcolor{lightgray}} 99.998 \% & {\cellcolor{lightgray}} 99.998 \% & {\cellcolor{lightgray}} 99.998 \% & {\cellcolor{lightgray}} 99.883 \% & {\cellcolor{lightgray}} 99.971 \% & {\cellcolor{lightgray}} 100.000 \% & {\cellcolor{lightgray}} 99.852 \% & {\cellcolor{lightgray}} 99.871 \% & {\cellcolor{lightgray}} 99.999 \% \\
\multirow[c]{2}{*}{Generator reactive power bounds slack} & 100.000 \% & 100.000 \% & 100.000 \% & 100.000 \% & 100.000 \% & 100.000 \% & 48.320 \% & 45.136 \% & 100.000 \% \\
 & {\cellcolor{lightgray}} 100.000 \% & {\cellcolor{lightgray}} 100.000 \% & {\cellcolor{lightgray}} 99.871 \% & {\cellcolor{lightgray}} 99.911 \% & {\cellcolor{lightgray}} 99.913 \% & {\cellcolor{lightgray}} 99.860 \% & {\cellcolor{lightgray}} 45.374 \% & {\cellcolor{lightgray}} 44.625 \% & {\cellcolor{lightgray}} 99.966 \% \\
\multirow[c]{2}{*}{Generator real power bounds slack} & 51.656 \% & 34.983 \% & 0.000 \% & 47.628 \% & 44.315 \% & 0.000 \% & 63.454 \% & 65.336 \% & 0.000 \% \\
 & {\cellcolor{lightgray}} 64.226 \% & {\cellcolor{lightgray}} 52.413 \% & {\cellcolor{lightgray}} 0.000 \% & {\cellcolor{lightgray}} 49.203 \% & {\cellcolor{lightgray}} 47.558 \% & {\cellcolor{lightgray}} 0.000 \% & {\cellcolor{lightgray}} 65.456 \% & {\cellcolor{lightgray}} 65.212 \% & {\cellcolor{lightgray}} 0.000 \% \\
\hline
\end{tabular}
}
\caption{Post-power flow percentage of entities \textbf{below} $\mathbf{0.0001}$ constraint violation. The non-shaded rows are \textsc{FullTop} results, while the shaded rows are \textsc{TopDrop} results.}
\label{tab:post_0.0001}
\end{table}

\begin{table}[!h]
\centering
\resizebox{\textwidth}{!}{\begin{tabular}{lrrr|rrr|rrr}
\hline
 & \multicolumn{3}{c|}{500} & \multicolumn{3}{c|}{2000} & \multicolumn{3}{c}{10000} \\
 & \textsc{Wide-CANOS} & \textsc{Deep-CANOS}$_{48}$ & \textsc{DC-IPOPT} & \textsc{Deep-CANOS}$_{48}$ & \textsc{Deep-CANOS}$_{60}$ & \textsc{DC-IPOPT} & \textsc{Deep-CANOS}$_{48}$ & \textsc{Deep-CANOS}$_{60}$ & \textsc{DC-IPOPT} \\
\hline
\multirow[c]{2}{*}{Branch thermal limit from} & 99.974 \% & 99.981 \% & 99.846 \% & 99.989 \% & 99.988 \% & 99.807 \% & 99.987 \% & 99.988 \% & 99.985 \% \\
 & {\cellcolor{lightgray}} 99.941 \% & {\cellcolor{lightgray}} 99.977 \% & {\cellcolor{lightgray}} 99.843 \% & {\cellcolor{lightgray}} 99.988 \% & {\cellcolor{lightgray}} 99.988 \% & {\cellcolor{lightgray}} 99.808 \% & {\cellcolor{lightgray}} 99.988 \% & {\cellcolor{lightgray}} 99.989 \% & {\cellcolor{lightgray}} 99.985 \% \\
\multirow[c]{2}{*}{Branch thermal limit to} & 99.999 \% & 99.999 \% & 99.853 \% & 99.986 \% & 99.985 \% & 99.820 \% & 99.992 \% & 99.992 \% & 99.985 \% \\
 & {\cellcolor{lightgray}} 99.984 \% & {\cellcolor{lightgray}} 99.995 \% & {\cellcolor{lightgray}} 99.850 \% & {\cellcolor{lightgray}} 99.985 \% & {\cellcolor{lightgray}} 99.985 \% & {\cellcolor{lightgray}} 99.819 \% & {\cellcolor{lightgray}} 99.992 \% & {\cellcolor{lightgray}} 99.991 \% & {\cellcolor{lightgray}} 99.985 \% \\
\multirow[c]{2}{*}{Branch voltage angle difference} & 100.000 \% & 100.000 \% & 100.000 \% & 100.000 \% & 100.000 \% & 100.000 \% & 100.000 \% & 100.000 \% & 100.000 \% \\
 & {\cellcolor{lightgray}} 100.000 \% & {\cellcolor{lightgray}} 100.000 \% & {\cellcolor{lightgray}} 100.000 \% & {\cellcolor{lightgray}} 100.000 \% & {\cellcolor{lightgray}} 100.000 \% & {\cellcolor{lightgray}} 100.000 \% & {\cellcolor{lightgray}} 100.000 \% & {\cellcolor{lightgray}} 100.000 \% & {\cellcolor{lightgray}} 100.000 \% \\
\multirow[c]{2}{*}{Bus voltage bounds pq} & 99.194 \% & 99.505 \% & 100.000 \% & 99.702 \% & 99.767 \% & 100.000 \% & 99.729 \% & 99.762 \% & 99.597 \% \\
 & {\cellcolor{lightgray}} 98.914 \% & {\cellcolor{lightgray}} 99.141 \% & {\cellcolor{lightgray}} 99.992 \% & {\cellcolor{lightgray}} 99.669 \% & {\cellcolor{lightgray}} 99.756 \% & {\cellcolor{lightgray}} 99.999 \% & {\cellcolor{lightgray}} 99.722 \% & {\cellcolor{lightgray}} 99.735 \% & {\cellcolor{lightgray}} 99.599 \% \\
\multirow[c]{2}{*}{Bus voltage bounds pv} & 100.000 \% & 100.000 \% & 100.000 \% & 99.921 \% & 99.985 \% & 100.000 \% & 99.807 \% & 99.827 \% & 100.000 \% \\
 & {\cellcolor{lightgray}} 99.998 \% & {\cellcolor{lightgray}} 99.998 \% & {\cellcolor{lightgray}} 99.998 \% & {\cellcolor{lightgray}} 99.881 \% & {\cellcolor{lightgray}} 99.969 \% & {\cellcolor{lightgray}} 100.000 \% & {\cellcolor{lightgray}} 99.783 \% & {\cellcolor{lightgray}} 99.805 \% & {\cellcolor{lightgray}} 99.999 \% \\
\multirow[c]{2}{*}{Generator reactive power bounds slack} & 100.000 \% & 100.000 \% & 100.000 \% & 100.000 \% & 100.000 \% & 100.000 \% & 48.302 \% & 45.116 \% & 100.000 \% \\
 & {\cellcolor{lightgray}} 100.000 \% & {\cellcolor{lightgray}} 100.000 \% & {\cellcolor{lightgray}} 99.871 \% & {\cellcolor{lightgray}} 99.911 \% & {\cellcolor{lightgray}} 99.913 \% & {\cellcolor{lightgray}} 99.860 \% & {\cellcolor{lightgray}} 45.349 \% & {\cellcolor{lightgray}} 44.598 \% & {\cellcolor{lightgray}} 99.966 \% \\
\multirow[c]{2}{*}{Generator real power bounds slack} & 51.629 \% & 34.960 \% & 0.000 \% & 47.624 \% & 44.309 \% & 0.000 \% & 63.454 \% & 65.336 \% & 0.000 \% \\
 & {\cellcolor{lightgray}} 64.224 \% & {\cellcolor{lightgray}} 52.404 \% & {\cellcolor{lightgray}} 0.000 \% & {\cellcolor{lightgray}} 49.200 \% & {\cellcolor{lightgray}} 47.556 \% & {\cellcolor{lightgray}} 0.000 \% & {\cellcolor{lightgray}} 65.453 \% & {\cellcolor{lightgray}} 65.212 \% & {\cellcolor{lightgray}} 0.000 \% \\
\hline
\end{tabular}
}
\caption{Post-power flow percentage of entities below \textbf{below} $\mathbf{10^{-6}}$ constraint violation. The non-shaded rows are \textsc{FullTop} results, while the shaded rows are \textsc{TopDrop} results.}
\label{tab:post_1e-06}
\end{table}

\begin{table}[!h]
\centering
\resizebox{\textwidth}{!}{\begin{tabular}{lrrr|rrr|rrr}
\hline
 & \multicolumn{3}{c|}{500} & \multicolumn{3}{c|}{2000} & \multicolumn{3}{c}{10000} \\
 & \textsc{Wide-CANOS} & \textsc{Deep-CANOS}$_{48}$ & \textsc{DC-IPOPT} & \textsc{Deep-CANOS}$_{48}$ & \textsc{Deep-CANOS}$_{60}$ & \textsc{DC-IPOPT} & \textsc{Deep-CANOS}$_{48}$ & \textsc{Deep-CANOS}$_{60}$ & \textsc{DC-IPOPT} \\
\hline
\multirow[c]{2}{*}{Branch thermal limit from} & 99.974 \% & 99.981 \% & 99.846 \% & 99.989 \% & 99.988 \% & 99.807 \% & 99.987 \% & 99.988 \% & 99.985 \% \\
 & {\cellcolor{lightgray}} 99.941 \% & {\cellcolor{lightgray}} 99.977 \% & {\cellcolor{lightgray}} 99.843 \% & {\cellcolor{lightgray}} 99.988 \% & {\cellcolor{lightgray}} 99.988 \% & {\cellcolor{lightgray}} 99.808 \% & {\cellcolor{lightgray}} 99.988 \% & {\cellcolor{lightgray}} 99.989 \% & {\cellcolor{lightgray}} 99.985 \% \\
\multirow[c]{2}{*}{Branch thermal limit to} & 99.999 \% & 99.999 \% & 99.853 \% & 99.986 \% & 99.985 \% & 99.820 \% & 99.992 \% & 99.992 \% & 99.985 \% \\
 & {\cellcolor{lightgray}} 99.984 \% & {\cellcolor{lightgray}} 99.995 \% & {\cellcolor{lightgray}} 99.850 \% & {\cellcolor{lightgray}} 99.985 \% & {\cellcolor{lightgray}} 99.985 \% & {\cellcolor{lightgray}} 99.819 \% & {\cellcolor{lightgray}} 99.992 \% & {\cellcolor{lightgray}} 99.991 \% & {\cellcolor{lightgray}} 99.985 \% \\
\multirow[c]{2}{*}{Branch voltage angle difference} & 100.000 \% & 100.000 \% & 100.000 \% & 100.000 \% & 100.000 \% & 100.000 \% & 100.000 \% & 100.000 \% & 100.000 \% \\
 & {\cellcolor{lightgray}} 100.000 \% & {\cellcolor{lightgray}} 100.000 \% & {\cellcolor{lightgray}} 100.000 \% & {\cellcolor{lightgray}} 100.000 \% & {\cellcolor{lightgray}} 100.000 \% & {\cellcolor{lightgray}} 100.000 \% & {\cellcolor{lightgray}} 100.000 \% & {\cellcolor{lightgray}} 100.000 \% & {\cellcolor{lightgray}} 100.000 \% \\
\multirow[c]{2}{*}{Bus voltage bounds pq} & 99.189 \% & 99.499 \% & 100.000 \% & 99.702 \% & 99.767 \% & 100.000 \% & 99.729 \% & 99.761 \% & 99.597 \% \\
 & {\cellcolor{lightgray}} 98.911 \% & {\cellcolor{lightgray}} 99.138 \% & {\cellcolor{lightgray}} 99.992 \% & {\cellcolor{lightgray}} 99.669 \% & {\cellcolor{lightgray}} 99.756 \% & {\cellcolor{lightgray}} 99.999 \% & {\cellcolor{lightgray}} 99.721 \% & {\cellcolor{lightgray}} 99.735 \% & {\cellcolor{lightgray}} 99.599 \% \\
\multirow[c]{2}{*}{Bus voltage bounds pv} & 100.000 \% & 100.000 \% & 100.000 \% & 99.921 \% & 99.985 \% & 100.000 \% & 99.806 \% & 99.827 \% & 100.000 \% \\
 & {\cellcolor{lightgray}} 99.998 \% & {\cellcolor{lightgray}} 99.998 \% & {\cellcolor{lightgray}} 99.998 \% & {\cellcolor{lightgray}} 99.881 \% & {\cellcolor{lightgray}} 99.969 \% & {\cellcolor{lightgray}} 100.000 \% & {\cellcolor{lightgray}} 99.782 \% & {\cellcolor{lightgray}} 99.804 \% & {\cellcolor{lightgray}} 99.999 \% \\
\multirow[c]{2}{*}{Generator reactive power bounds slack} & 100.000 \% & 100.000 \% & 100.000 \% & 100.000 \% & 100.000 \% & 100.000 \% & 48.302 \% & 45.116 \% & 100.000 \% \\
 & {\cellcolor{lightgray}} 100.000 \% & {\cellcolor{lightgray}} 100.000 \% & {\cellcolor{lightgray}} 99.871 \% & {\cellcolor{lightgray}} 99.911 \% & {\cellcolor{lightgray}} 99.913 \% & {\cellcolor{lightgray}} 99.860 \% & {\cellcolor{lightgray}} 45.349 \% & {\cellcolor{lightgray}} 44.598 \% & {\cellcolor{lightgray}} 99.966 \% \\
\multirow[c]{2}{*}{Generator real power bounds slack} & 51.629 \% & 34.960 \% & 0.000 \% & 47.624 \% & 44.309 \% & 0.000 \% & 63.454 \% & 65.336 \% & 0.000 \% \\
 & {\cellcolor{lightgray}} 64.224 \% & {\cellcolor{lightgray}} 52.404 \% & {\cellcolor{lightgray}} 0.000 \% & {\cellcolor{lightgray}} 49.200 \% & {\cellcolor{lightgray}} 47.556 \% & {\cellcolor{lightgray}} 0.000 \% & {\cellcolor{lightgray}} 65.453 \% & {\cellcolor{lightgray}} 65.212 \% & {\cellcolor{lightgray}} 0.000 \% \\
\hline

\end{tabular}
}
\caption{Post-power flow percentage of entities \textbf{below} $\mathbf{10^{-8}}$ constraint violation, for \textsc{DC-IPOPT} and  \textsc{CANOS} across different grid sizes. The non-shaded rows are \textsc{FullTop} results, while the shaded rows are \textsc{TopDrop} results.}
\label{tab:post_1e-08}
\end{table}

\clearpage
\onecolumn
\section{Model Size Analysis}
\label{apdx:model_size_analysis}

\begin{figure}[!h]
\centering
\begin{subfigure}[b]{0.45\textwidth}
  \centering
    \includegraphics[width=\textwidth]{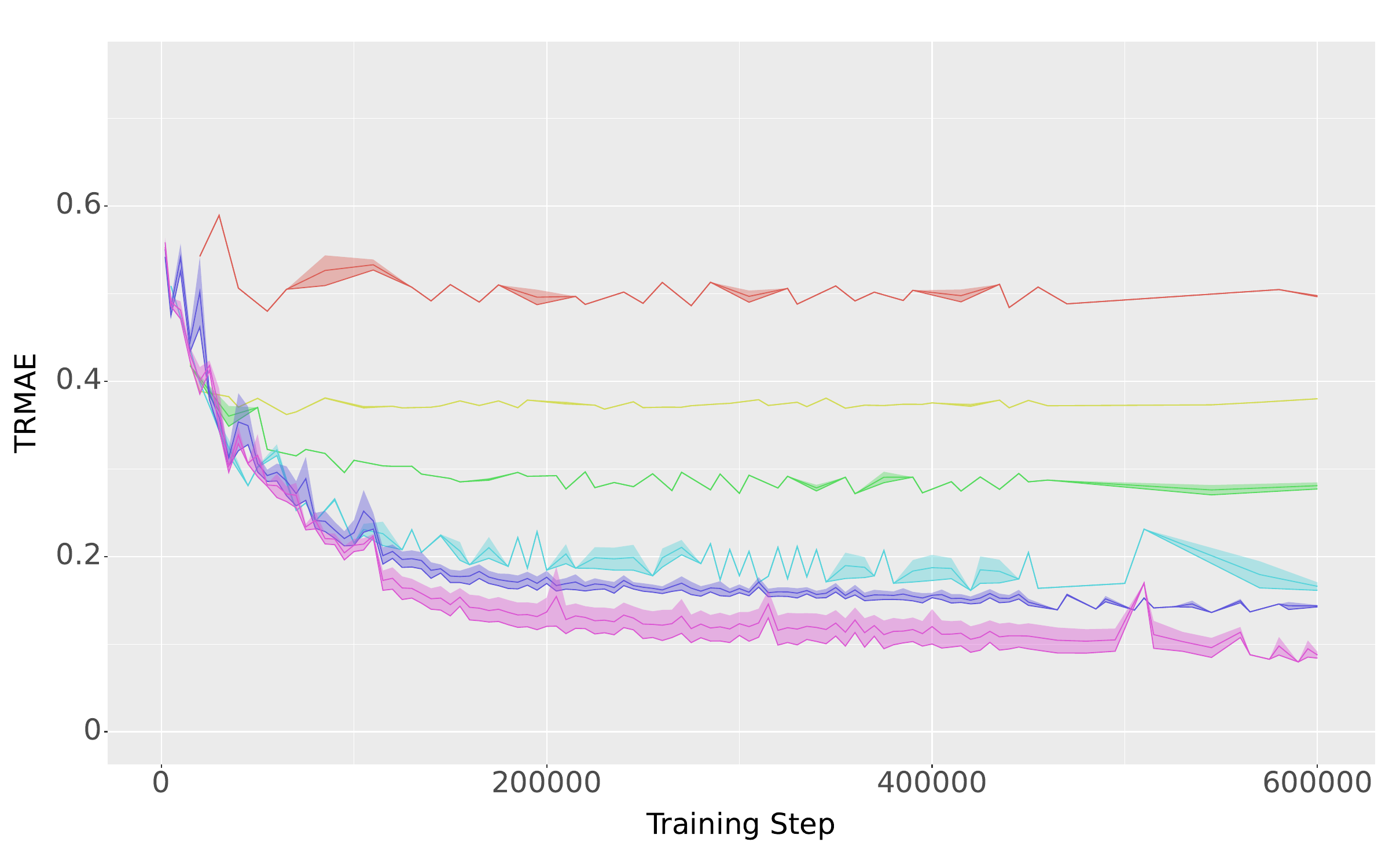}
    %\caption{TRMAE Decreases With More Message Passing Steps}
    %\label{fig:sweep_steps_rmae}
\end{subfigure}
\begin{subfigure}[b]{0.45\textwidth}
  \centering
  \includegraphics[width=\textwidth]{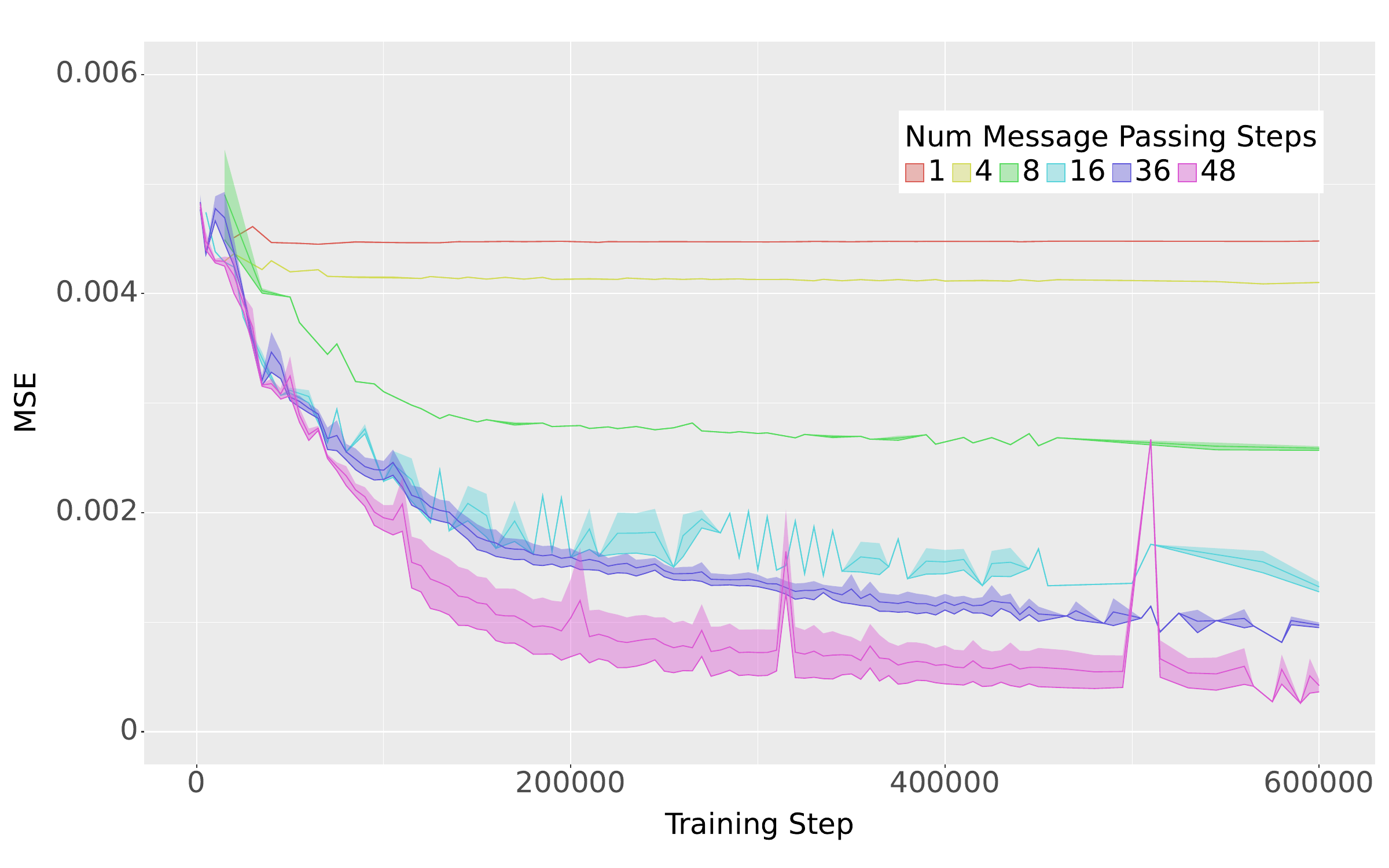}
  %\caption{MSE Decreases With More Message Passing Steps}
  %\label{fig:sweep_steps_mse}
\end{subfigure}
\caption{\textbf{Supervised performance as a function of Number of Message Passing Steps:} Both the TRMAE (left) and MSE (right) decrease as the number of message passing steps (and consequently number of parameters) increases in the model. These metrics on the validation split of the data throughout training.}
\label{fig:sweep_steps_supervised}
\end{figure}

\begin{figure}[!h]
\centering
  \centering
    \includegraphics[width=\textwidth]{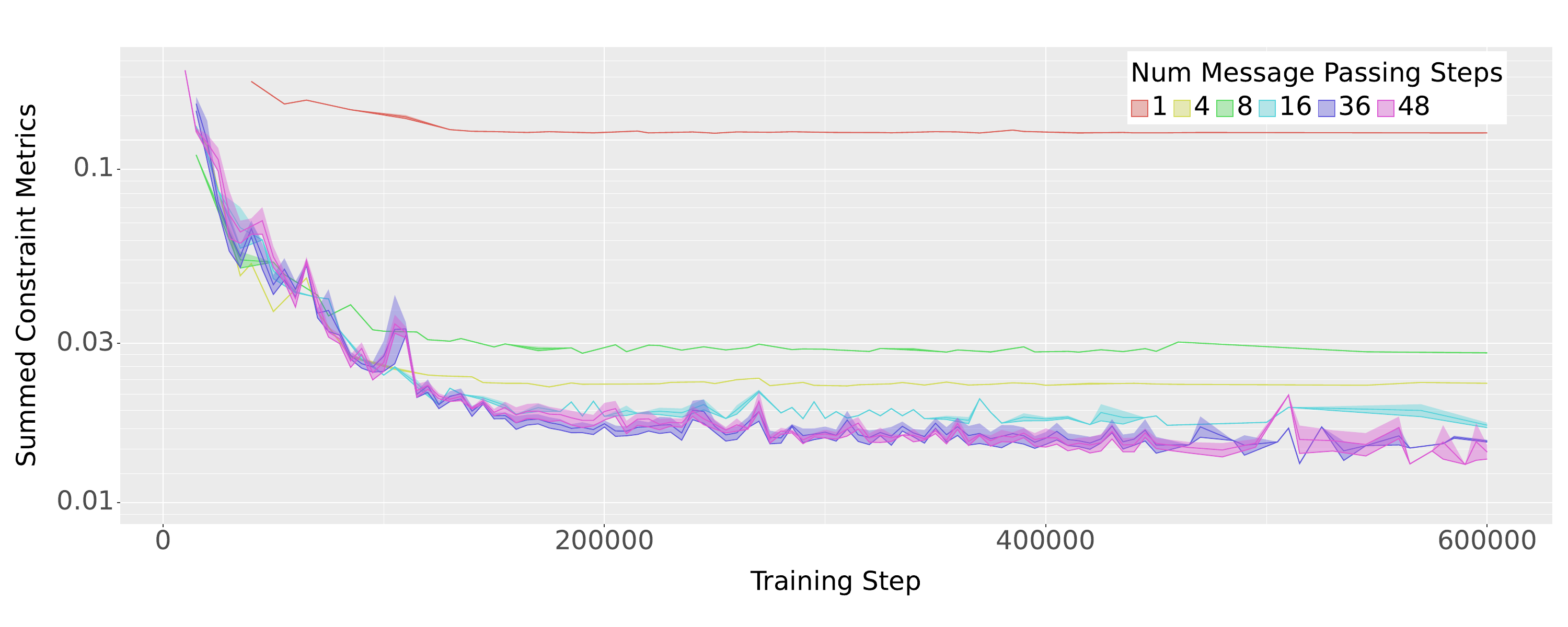}
    \caption{\textbf{Feasibility as a function of Number of Message Passing Steps:} Feasibility substantially improves as we increase the number of message passing steps.}
    \label{fig:sweep_steps_constraints}
\end{figure}

In the main text, we report results for several \textsc{CANOS} variants. The main hyperparameter across these models is the number of message-passing steps, where each message-passing step is a InteractionNetwork block with its own unique set of parameters. This affects the model in two ways. First, models with more message-passing steps can propagate local information further along the graph. Second, models with more message-passing steps have a larger number of parameters. Before collecting our main results, we conducted an analysis on the impact of the number of message-passing steps on model performance before power flow. We conducted this on the 500-bus grid with the \textsc{TopDrop} dataset. We used two random seeds for each model size and report the mean and 95\% confidence interval in the figures below.  Our analyses reveal large improvements in supervised performance (Figure \ref{fig:sweep_steps_supervised}) and feasibility (Figure \ref{fig:sweep_steps_constraints}) as we increase the model size. Thus, its evident that a key aspect of our superior model performance is due to using a large number of message passing steps. We saw that this effect was exacerbated on larger grids, which is why we switched to using the \textsc{CANOS} variants with more message passing steps (i.e. 48, 60) on the 2000- and 10000- bus grids.

% FALCONER: use Graph Convolution Layers with 2 or 3 

\end{document}